\documentclass{article}

\usepackage[final]{neurips_2022}

\usepackage[utf8]{inputenc} 
\usepackage[T1]{fontenc}    
\usepackage{hyperref}       
\usepackage{url}            
\usepackage{booktabs}       
\usepackage{amsfonts}       
\usepackage{nicefrac}       
\usepackage{microtype}      
\usepackage[dvipsnames]{xcolor}

\usepackage{pifont}
\usepackage{amsmath}
\usepackage{amssymb}
\usepackage{wrapfig}
\usepackage{multirow}
\usepackage{multicol}
\usepackage{tabularx}
\usepackage{graphicx}
\usepackage{xcolor}
\usepackage{soul}
\usepackage{xspace}
\usepackage{arydshln}
\usepackage{caption}
\usepackage{nicematrix}
\usepackage{wrapfig}
\usepackage{paralist}
\usepackage{enumerate}
\usepackage{enumitem}

\usepackage[compact]{titlesec}
\titlespacing{\section}{0pt}{2ex}{1ex}
\titlespacing{\subsection}{0pt}{1ex}{0ex}
\titlespacing{\subsubsection}{0pt}{0.5ex}{0ex}
    
\newcommand{\methodname}{\textsc{NaturalProver} }
\newcommand{\methodnamenospace}{\textsc{NaturalProver}\xspace}
\newcommand{\mnamepp}{$\textsc{NaturalProver}_{++}$\xspace}

\newcommand{\xb}{\ensuremath{\mathbf{x}}}

\newcommand{\yb}{\ensuremath{\mathbf{y}}}
\newcommand{\rb}{\ensuremath{\mathbf{r}}}

\newcommand{\natproofs}{\ensuremath{\textsc{NaturalProofs}}}
\DeclareMathOperator*{\argmax}{arg\,max}

\DeclareMathOperator*{\argtopK}{arg\,\mathrm{top}\text{-}\mathrm{K}}
\newcommand{\cmark}{\ding{51}}%
\newcommand{\xmark}{\ding{55}}%

\definecolor{colorOverallCorrect}{HTML}{C6EFCA}
\definecolor{colorOverallUseful}{HTML}{CBEEFE}
\definecolor{colorStepCorrect}{HTML}{C6EFCA}
\definecolor{colorStepUseful}{HTML}{CBEEFE}
\definecolor{colorRefErrs}{HTML}{F6C8C4}
\definecolor{colorDeploy}{HTML}{FFFFFF}
\definecolor{colorJustify}{HTML}{FFFFFF}
\definecolor{colorHalluc}{HTML}{FFFFFF}
\definecolor{colorLoop}{HTML}{FFFFFF}
\definecolor{colorEqnErrs}{HTML}{FFF2CD}
\definecolor{colorEquation}{HTML}{FFFFFF}
\definecolor{colorDerivation}{HTML}{FFFFFF}
\definecolor{colorOtherErrs}{HTML}{FFD5B2}
\definecolor{colorSkip}{HTML}{FFFFFF}
\definecolor{colorRepetition}{HTML}{FFFFFF}
\definecolor{colorInvalid}{HTML}{FFFFFF}
\definecolor{colorLangErrs}{HTML}{E6E1D2}
\definecolor{colorIncomplete}{HTML}{FFFFFF}
\definecolor{colorMisformat}{HTML}{FFFFFF}
\definecolor{colorUnk}{HTML}{FFFFFF}
\definecolor{colorSymErrs}{HTML}{D0CEC4}
\definecolor{colorUndefined}{HTML}{FFFFFF}
\definecolor{colorOverloaded}{HTML}{FFFFFF}
\definecolor{colorMistyped}{HTML}{FFFFFF}
\definecolor{colorUnconventional}{HTML}{FFFFFF}
\definecolor{ref}{HTML}{FFFFFF}
\definecolor{ref2}{HTML}{FFFFFF}
\definecolor{ref3}{HTML}{779CDB}

\newcommand{\cmdOverallCorrect[1]}{\sethlcolor{colorOverallCorrect}{\hl{~#1~}}}
\newcommand{\cmdOverallUseful[1]}{\sethlcolor{colorOverallUseful}{{\hl{~#1~}}}}
\newcommand{\cmdStepCorrect[1]}{\sethlcolor{colorStepCorrect}{{\hl{~#1~}}}}
\newcommand{\cmdStepUseful[1]}{\sethlcolor{colorStepUseful}{{\hl{~#1~}}}}
\newcommand{\cmdRefErrs[1]}{\sethlcolor{colorRefErrs}{{\hl{~#1~}}}}
\newcommand{\cmdDeploy[1]}{\sethlcolor{colorDeploy}{{\hl{~#1~}}}}
\newcommand{\cmdJustify[1]}{\sethlcolor{colorJustify}{{\hl{~#1~}}}}
\newcommand{\cmdHalluc[1]}{\sethlcolor{colorHalluc}{{\hl{~#1~}}}}
\newcommand{\cmdLoop[1]}{\sethlcolor{colorLoop}{{\hl{~#1~}}}}
\newcommand{\cmdEqnErrs[1]}{\sethlcolor{colorEqnErrs}{{\hl{~#1~}}}}
\newcommand{\cmdEquation[1]}{\sethlcolor{colorEquation}{{\hl{~#1~}}}}
\newcommand{\cmdDerivation[1]}{\sethlcolor{colorDerivation}{{\hl{~#1~}}}}
\newcommand{\cmdOtherErrs[1]}{\sethlcolor{colorOtherErrs}{{\hl{~#1~}}}}
\newcommand{\cmdSkip[1]}{\sethlcolor{colorSkip}{{\hl{~#1~}}}}
\newcommand{\cmdRepetition[1]}{\sethlcolor{colorRepetition}{{\hl{~#1~}}}}
\newcommand{\cmdInvalid[1]}{\sethlcolor{colorInvalid}{{\hl{~#1~}}}}
\newcommand{\cmdLangErrs[1]}{\sethlcolor{colorLangErrs}{{\hl{~#1~}}}}
\newcommand{\cmdIncomplete[1]}{\sethlcolor{colorIncomplete}{{\hl{~#1~}}}}
\newcommand{\cmdMisformat[1]}{\sethlcolor{colorMisformat}{{\hl{~#1~}}}}
\newcommand{\cmdUnk[1]}{\sethlcolor{colorUnk}{{\hl{~#1~}}}}
\newcommand{\cmdSymErrs[1]}{\sethlcolor{colorSymErrs}{{\hl{~#1~}}}}
\newcommand{\cmdUndefined[1]}{\sethlcolor{colorUndefined}{{\hl{~#1~}}}}
\newcommand{\cmdOverloaded[1]}{\sethlcolor{colorOverloaded}{{\hl{~#1~}}}}
\newcommand{\cmdMistyped[1]}{\sethlcolor{colorMistyped}{{\hl{~#1~}}}}
\newcommand{\cmdUnconventional[1]}{\sethlcolor{colorUnconventional}{{\hl{~#1~}}}}
\newcommand{\cmdref[1]}{\sethlcolor{ref}{{\hl{~#1~}}}}
\newcommand{\cmdreff[1]}{\sethlcolor{ref2}{{\hl{~#1~}}}}

\title{\methodnamenospace: Grounded Mathematical Proof Generation with Language Models}

\author{%
  Sean Welleck$^{1,2*}$,
  Jiacheng Liu$^{1*}$,
  Ximing Lu$^{2}$,
  Hannaneh Hajishirzi$^{1,2}$,
  Yejin Choi$^{1,2}$ \\
  $^1$Paul G. Allen School of Computer Science \& Engineering, University of Washington \\
  $^2$Allen Institute for Artificial Intelligence, $^*$Equal contribution \\
  \texttt{wellecks@uw.edu}
}

\begin{document}

\maketitle

\begin{abstract}
Theorem proving in natural mathematical language -- 
the mixture of symbolic and natural language used by humans -- plays a central role in mathematical advances and education, and tests aspects of reasoning that are core to intelligence.
Yet it has remained underexplored with modern generative models.
We study large-scale language models on two new generation tasks:
suggesting the next step in a mathematical proof, and full proof generation.
We develop \methodnamenospace, a language model that generates proofs by conditioning on background references (e.g. theorems and definitions that are either retrieved or human-provided), and optionally enforces their presence with constrained decoding.
On theorems from the $\natproofs$ benchmark,
\methodnamenospace improves the quality of next-step suggestions and generated proofs over fine-tuned GPT-3, according to human evaluations from university-level mathematics students.
\methodnamenospace is capable of proving some theorems that require short (2-6 step) proofs, and providing next-step suggestions that are rated as correct and useful over 40\% of the time, which is to our knowledge the first demonstration of these capabilities using neural language models.\footnote{Code and data available at \url{https://github.com/wellecks/naturalprover}.}
\end{abstract}

\begin{figure}[!h]
    \centering
    \includegraphics[width=0.95\textwidth]{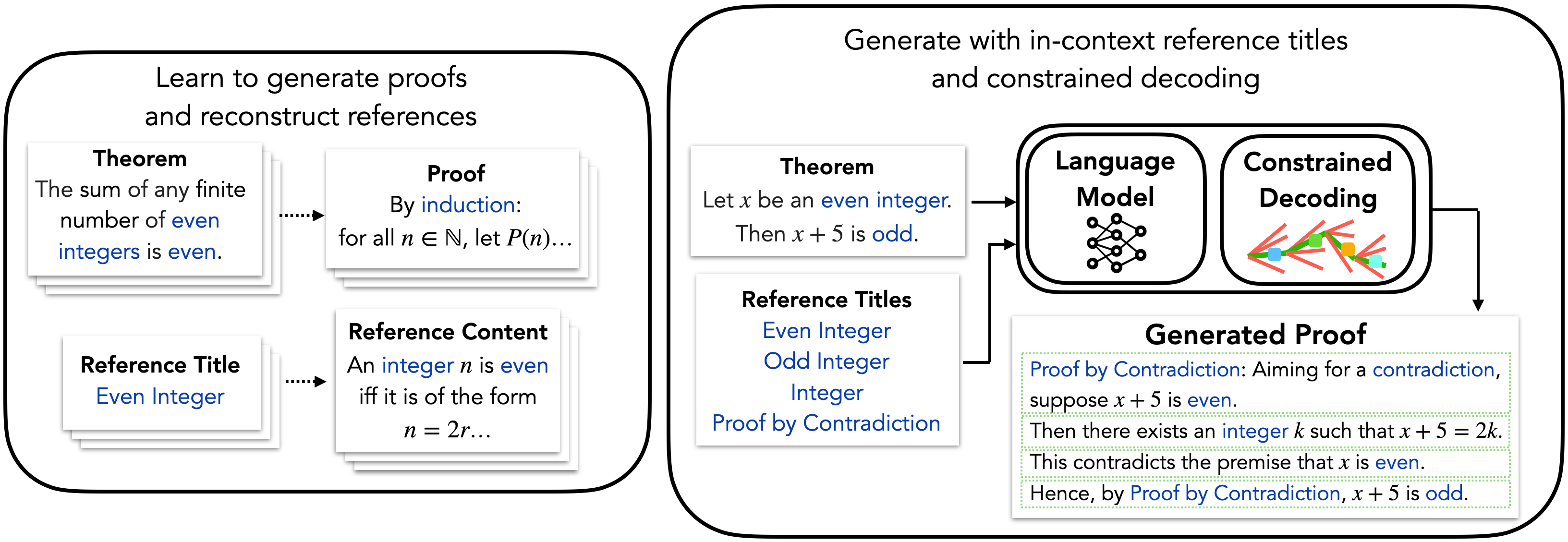}
    \caption{\methodname proves \href{https://proofwiki.org/wiki/Even_Integer_Plus_5_is_Odd}{Even Integer Plus 5 is Odd}.
    At training time, \methodname obtains background knowledge about references (e.g. theorems or definitions) via \textit{reference reconstruction}: learning to map a reference's title to its content.
    At test time, \methodname grounds its generations through in-context reference constraints that are retrieved or human-provided, and optionally enforced with \textit{stepwise constrained decoding}.
    This theorem's \href{https://proofwiki.org/wiki/Even_Integer_Plus_5_is_Odd/Proof_by_Contradiction}{human-written proof in ProofWiki} contains an error and differs substantially from \methodnamenospace's correct proof.
    }
    \label{fig:fig1}
\end{figure}

\section{Introduction}
Constructing a rational argument that justifies a claim is a key aspect of explaining, verifying, 
and communicating ideas in situations ranging from everyday interactions, to legal and political discourse, to science and mathematics~\citep{Davis1981,Voss1991,Kaye1992}.
Within the latter context, a \textit{mathematical proof} -- a sequence of logical arguments expressed in a mixture of symbolic and natural language -- assumes this role by providing justification  and insight into why a  claim is true~\citep{DeVilliers1990}. 
Proofs operate on a relatively  explicit and objective set of ground knowledge, isolating a subset of reasoning that is desirable for models that form the foundation of machine learning systems~\citep{Bommasani2021OnTO}.
Moreover, we envision assistive systems that provide suggested proofs or next-steps, analogous to language-model-based code suggestions (e.g. \hyperlink{https://copilot.github.com/}{GitHub CoPilot}~\citep{chen2021evaluating}) or formal proof assistants (e.g. \hyperlink{https://github.com/jesse-michael-han/lean-gptf}{GPT-$f$}~\citep{han2021proof}), which could make learning or using mathematics more productive and accessible.

To this end,
we study the capabilities of large-scale language models (e.g. GPT-3~\cite{brown2020gpt3}) on two new
theorem proving tasks in 
natural mathematical language:
\textit{next-step suggestion}, in which a model suggests the next step of a proof, and \textit{full-proof generation}, in which a model fully proves a claim.
As proofs are grounded in knowledge from past results (e.g. theorems, definitions), analogous to facts deployed in a conversation~\citep{dinan2018wizard}, prior rulings used in a legal opinion~\citep{jensen2014thinking}, or articles used to justify an answer~\citep{nakano2021WebGPT}, we develop a methodology for obtaining and using background knowledge to prove theorems with a generic language model.

We develop \methodnamenospace, a language model that generates proofs by conditioning on background references (e.g. theorems and definitions that are either retrieved or human-provided), and optionally enforces their presence with a constrained decoding algorithm that leverages the multi-step structure of proofs.
On a collection of theorems from the $\natproofs$ benchmark~\citep{welleck2021naturalproofs}, \methodnamenospace improves the quality of next-step suggestions and generated proofs over fine-tuned GPT-3, according to human evaluations from university-level mathematics students.
\methodnamenospace is capable of proving some theorems that require short (2-6 step) proofs, and providing next-step suggestions that are rated as correct and useful more than 40\% of the time, which is to our knowledge the first demonstration of these capabilities using neural language models.

Along with these successes, we study deficiencies in our current models.
We find that models can struggle with logical coherence on longer proofs, 
with providing valid justifications, and with performing multi-step symbolic derivations.
Taken together, 
our tasks, methodology, and evaluation 
show the feasibility of language models as interactive aids in mathematics, along with open challenges.
\section{\textsc{NaturalProofs-Gen} Dataset and Tasks}
\label{sec:task}

We create a $\textsc{NaturalProofs-Gen}$ dataset adapted from \natproofs~\citep{welleck2021naturalproofs}, and use the dataset for two tasks: suggesting the next step of a proof, and fully proving a theorem.

\textbf{\textsc{NaturalProofs-Gen}.}
$\textsc{NaturalProofs-Gen}$ adapts data  from  \natproofs, which contains theorem statements, proofs, definitions, and additional pages (e.g. axioms, corollaries) sourced  from \href{https://proofwiki.org/}{ProofWiki}, an online compendium of community-contributed mathematical proofs.
In $\textsc{NaturalProofs-Gen}$, each example $(\xb, \yb) \in \mathcal{D}$ pairs a theorem $\xb$ with a gold proof $\yb$, both of which are a mixture of text and \LaTeX.
\cite{welleck2021naturalproofs} split the examples and reference sets into training, dev, and test splits to ensure that no theorem in the dev or test splits was mentioned in the training split.
We adopt these splits of roughly 12.5k training, 1k validation, and 1k test examples, and sampled \textit{core evaluation sets} with 100 dev and 100 test theorems that are used for human evaluation.
The proofs contain  additional structure, discussed next.

\textbf{Multi-step proof structure.}
Each proof has a \textit{multi-step} structure, meaning that a proof $\yb=(y_1,\ldots,y_{|\yb|})$ is a variable-length token sequence that is segmented into \textit{proof steps}, where each step $y_t$ is itself a variable-length sequence of tokens (either text or Latex).
The segmentation is largely determined by ProofWiki's formatting and community standards for structuring proofs, and we additionally merge steps to ensure that each step contains non-trivial semantic content.
For example, Figure~\ref{fig:fig1} shows a 4-step (generated) proof with each step highlighted in green.

\textbf{References.}
Each proof mentions a variable-number of \textit{references} $\{\rb_1,\ldots,\rb_{R_y}\}$ from a set $\mathcal{R}$ of roughly 33k theorems and definitions, analogous to how Wikipedia articles reference other pages.
For example, Figure~\ref{fig:fig1} shows a proof with reference mentions in blue.
Each mention identifies a reference by its title and provides a natural language surface form.
For instance, in \autoref{fig:fig1}, the first proof step mentions the definition of even integer as \textcolor{blue}{even}, which is formatted in the proof as \texttt{[[Definition:Even\_Integer|even]]} and tokenized along with the rest of the proof.

\paragraph{Tasks.} We consider two tasks that are motivated by an assistive system that provides suggested proofs or next-steps to a user.
The \textbf{full proof generation} task is to generate a proof $\yb$ given a theorem $\xb$.
The \textbf{next-step suggestion} task is to generate a set of next steps $\{y_t^{k}\}_{k=1}^K$ given theorem $\xb$ and proof history $y_{<t}$ from a gold proof.
In each case, we consider an additional \textbf{provided reference} setting where the model is also given the set of references $\{\rb_1^*,\ldots,\rb_{R_y}^*\}$ from a gold proof of the theorem.
The next-step task simulates a human correctly proving the theorem up to a point, then querying a system for suggested next-steps when stuck, while the provided reference setting simulates a human specifying a plan for a system that writes a proof.

\section{\methodnamenospace: Grounded Proof Generation via Language Modeling}
\label{sec:method}

We describe \methodnamenospace, a language model which generates grounded proofs by conditioning on  references and optionally enforcing their presence with constrained decoding.

\textbf{Setup.} Our objective is to generate correct proofs,
    $\hat\yb = \argmax_{\yb} \mathrm{correct}(\xb,\yb)$.
Unfortunately, evaluating proof correctness is costly, and is only done once at test time.
A naive approach is to approximate the objective,
$\hat\yb \approx \arg\max_{\yb} \log p_\theta(\yb|\xb)$, by fine-tuning a language model $p_\theta$ on $(\xb,\yb)$ examples and using a decoding algorithm (e.g. greedy decoding).
We instead investigate conditioning on background knowledge in the form of reference documents, $p_\theta(\yb|\xb,R)$, which is  beneficial in related generation settings (e.g. \cite{shuster-etal-2021-retrieval-augmentation}), and offers control over the generated proof.
To do so, \methodname uses in-context references and a reference reconstruction objective.

\textbf{In-context references.} Language models have a limited context window that prevents conditioning on  full documents. Instead,
\methodname conditions on a set of reference titles, 
$p_\theta(\yb|\xb,R_{\text{title}})$.
Concretely, we fine-tune on (theorem, reference titles, proof) sequences of the form,
\begin{align}
\label{eqn:ref-cond}
    \nonumber&\scriptsize\texttt{<theorem> <title> \{theorem-title\} </title> <content> \{theorem-content\} </content> </theorem>}\\
    &\scriptsize\texttt{<ref> \{ref-title-1\} </ref> ... <ref> \{ref-title-R\} </ref>}\quad 
    \scriptsize\texttt{<proof> \{proof\} </proof>}
\end{align}
with new-lines and $\{\}$ tokens omitted, relevant 
strings inserted, and loss only on tokens after {\scriptsize\texttt{<proof>}}.

\textbf{Reference reconstruction.}
Reference titles 
do not  capture all of the information contained in the reference documents.
We learn a mapping between each reference title and its underlying document with a reference reconstruction objective, $p_\theta(\rb|\rb_{\text{title}})$ for references $\rb$ in the training reference set.
Concretely, we fine-tune on additional (title, content) pairs of the form,
\begin{align}
\label{eqn:ref-rec}
    \scriptsize
    \texttt{<\{type\}> <title> \{title\} </title> <content> \{content\} </content> </\{type\}>},
\end{align}
where the $\texttt{\{type\}}$ is theorem/definition/other, and the loss is only on tokens after {\scriptsize{\texttt{<content>}}}.
Intuitively, this lets the model associate each reference title with the reference's underlying content.

\paragraph{The joint objective.}
For training, we minimize the joint loss,
\begin{align}
\mathcal{L}(\theta)
&= \frac{1}{|\mathcal{D}^{\text{train}}| + |\mathcal{R}^{\text{train}}|} \Big[ \sum_{(\xb, \yb) \in \mathcal{D}^{\text{train}}}{-\log p_\theta(\yb | \xb, R_{\text{title}})} + \sum_{\rb \in \mathcal{R}^{\text{train}}}{-\log p_\theta(\rb | \rb_{\text{title}})} \Big].
\end{align}

\paragraph{Evaluation-time references.} We consider two settings for evaluation-time references: (i) \textit{retrieved} references, from a retrieval model $f(\xb)\rightarrow \{ \rb_1,\ldots,\rb_k \}$, and (ii) \textit{human-provided} references from the ground-truth proof.
The retrieval setting simulates a fully automated proof assistant, while the second 
simulates a human specifying a plan for an assistant that writes a proof, and 
acts as an upper bound for a retrieval system optimized to predict references in a ground-truth proof. 

\subsection{Stepwise constrained decoding}
\label{sec:decoding}

In the provided-reference setting, the conditioned references are known to be relevant to a correct proof.
We hypothesize that explicitly encouraging generated proofs to contain the references will improve correctness, by placing lexical constraints on the reference-titles at decoding time,
\begin{align}
\label{eqn:constrained}
    \hat\yb\approx &\argmax_{\yb} \log p_\theta(\yb|\xb,R_{\text{title}}),\ \text{subject to} \sum_{\rb_{\text{title}}\in R_{\text{title}}}\mathbb{I}\left[\rb_{\text{title}}\in \yb\right] = |R_{\text{title}}|,
\end{align}
where $\mathbb{I}\left[\cdot\right]$ is an indicator function.
To approximate this objective, we generate step-by-step by sampling multiple proof-step candidates, retaining those with high value (reference coverage and log-probability) in a beam, and continuing to the next step, which we call stepwise beam search.

\textbf{Value function.} The search supports any function of the proof-so-far, $v(y_{\leq t})\rightarrow \mathbb{R}$.
We use a value function that is a weighted combination of constraint satisfaction and log-probability, 
\begin{align}
    v_\alpha(y_{\leq t})=\alpha v_{\text{constraint}}(y_{\leq t}) + (1-\alpha)v_{\text{LM}}(y_{\leq t}),
\end{align}
where $v_{\text{constraint}}(y_{\leq t})$ is the number of unique in-context reference-titles in $y_{\leq t}$, and $v_{\text{LM}}(y_{\leq t})$ is $\log p_\theta(y_{\leq t})$. We normalize each term by dividing by the maximum absolute value among candidates.

\textbf{Stepwise beam search.} The procedure generates a proof $\yb=(y_1,\ldots,y_T)$ by iteratively sampling and pruning next-proof-step candidates $y_t$.
Each iteration expands a size-$K$ beam of proofs-so-far, $S_{t-1}=\{y_{< t}^{k}\}_{k=1}^K$, by generating $N$ next-step candidates,
\begin{align}\label{eqn:expand}
S_t'=\cup_{y_{<t}\in S_{t-1}}\big\{ (y_{<t} \circ y^{n}_{t})\ |\ y^n_t\sim q(\cdot |y_{< t},\xb,R_{\text{title}})\big\}_{n=1}^N,
\end{align}
where $q$ is a decoding algorithm (e.g. temperature sampling) and $\circ$ is concatenation. 
The next iteration's beam is formed by selecting the top scoring candidates,
    $S_{t}=\argtopK_{y_{\leq t}\in S_t'}\ v_\alpha(y_{\leq t})$.
When a proof in the beam terminates, it is not expanded further. The search ends when the beam consists of $K$ terminated proofs.
The highest value proof is returned as the final output.

\textbf{Stepwise++.} 
We add two mechanisms for promoting  exploration at each step.
First, we expand each prefix in the beam (Eqn. \ref{eqn:expand}) by sampling with multiple temperatures,
    $\{y_t^n\sim q_\tau(\cdot|y_{<t},\xb,R_{\text{title}})\ |\ \tau\in \{\tau_i\}_{i=1}^{m}\}$,
where $q_\tau$ is sampling with temperature $\tau$.
This relaxes the commitment to a single temperature for all proof steps, balancing exploration (higher $\tau$) with exploitation (lower $\tau$).

Second, rather than selecting the top-K candidates, we select clusters based on different value weights: $S_t=\cup_{\alpha\in \{\alpha_j\}_{j=1}^{\ell}}\mathrm{top}_{K'} (S_t^\alpha)$,
where $S_t^\alpha$ is the set of candidates scored with $v_\alpha$, and $K'=K/\ell$.
This interpolates between selecting steps based on likelihood (low $\alpha$) and constraint satisfaction (high $\alpha$).

\textbf{Full proof sampling and greedy decoding.} 
An alternative is to sample full proofs and select the best one according to the value function.
This can be viewed as expansion (Eqn. \ref{eqn:expand}) done at the full proof, rather than the step level.
Moreover, greedy decoding corresponds to expanding only 1 candidate with temperature $\rightarrow 0$.
We formalize this in \S\ref{sec:expr_decoding} as a segment-level search that contains stepwise++, full proof sampling, and greedy decoding as special cases.

\section{Proof Evaluation}
\label{sec:evaluation}

A proof's correctness is contingent on a variety of factors,
including reasoning with past results, performing symbolic derivations, and altogether providing sufficient evidence that the claim is true. 
We design a human-evaluation schema that isolates these aspects at the proof-step level, along with a full-proof summary.
\autoref{tab:schema_examples} summarizes the schema, which we overview below.

\textbf{\cmdRefErrs[References.]}
First, proofs involve deploying statements from references, such as applying a definition or adapting it to fit the context. 
Deployments should be consistent with the reference,
e.g. deploying the \hyperlink{https://proofwiki.org/wiki/Definition:Even_Integer}{definition of even integer} as `...by definition, $\exists k\in \mathbb{Z} : x=2k$...', rather than `...$\exists k\in \mathbb{Z} :x=2k+1$', and are a common source of errors in student proofs~\citep{edwards2004}.

\renewcommand{\arraystretch}{1.2} 
\begin{table*}[ht]
\centering
\small
\begin{tabularx}{\textwidth}{lX}
\toprule
\textbf{Error Type} & \textbf{Example} \\
\midrule
\cmdRefErrs[\bf Reasoning: Reference] \\ 
\hspace{1em} \cmdDeploy[Invalid Deployment] & Since $x$ is an \textcolor{black}{\underline{even integer}}, $\exists k \in \mathbb{Z}: x = 2 k + 1$. \\
\hspace{1em} \cmdJustify[Invalid Justification] & $\mathbb{E}(X^2) = \sum_{k=1}^{n}{k^2 \text{Pr}(X = k)}$ \quad \textcolor{black}{\underline{Power Series for Exponential Function}} \\ 
\hspace{1em} \cmdHalluc[Hallucinated Ref.] & From \underline{\cmdRefErrs[Power of Number are Irrational]}, $\sqrt[3]{2}$ is irrational. \\
\hspace{1em} \cmdLoop[Self Loop] & \textcolor{gray}{(Proving Pythagoras's Theorem:)} \quad From \textcolor{black}{\underline{Pythagoras's Theorem}}, $c^2 = a^2 + b^2$. \\
\midrule
\cmdEqnErrs[\bf Reasoning: Equation] & \\
\hspace{1em} \cmdEquation[Invalid Equation] & $\forall x \in \mathbb{R}, x + x = 3x$. \\
\hspace{1em} \cmdDerivation[Invalid Derivation] & \textcolor{gray}{(Since $x$ is an even integer, $x + 1 = 2 r + 1$)} \quad $ = 2 (r + 1)$ \\
\midrule
\cmdOtherErrs[\bf Reasoning: Other] & \\
\hspace{1em} \cmdSkip[Skips Steps] & \textcolor{gray}{($x \in \mathbb{Z}$ is not a multiple of 3.)} \quad Therefore, $x^3 \equiv 1 \text{ or } 8 (\text{mod } 9)$ \\
\hspace{1em} \cmdRepetition[Repetition] & \textcolor{gray}{(Let $\triangle{ABC}$ be a right triangle.)} \quad Then $\triangle{ABC}$ is a right triangle. \\
\hspace{1em} \cmdInvalid[Invalid (Other)] & \textcolor{gray}{($x$ is an even integer.)} \quad So, $x + 1$ is an even integer. \\
\midrule
\cmdLangErrs[\bf Language] & Let $c = \sqrt{a^2 \text{\textcolor{black}{ $\backslash{}$add }} b^2}$ be the \quad \textit{(\cmdIncomplete[incomplete statement]; \cmdUnk[unknown symbol ]$\backslash{}add$)} \\
\midrule
\cmdSymErrs[\bf Symbolic] & \textcolor{gray}{(Let $x \in \mathbb{R}$.)} \quad Let $y = x \textcolor{black}{\circ} x^{-1}$. \quad \textit{(\cmdUndefined[undefined operator $\circ$ for real numbers])} \\
\bottomrule
\end{tabularx}%
\caption{
    Overview of human evaluation error schema.
    See \autoref{tab:schema_definitions} for the full schema.
    \textcolor{black}{\underline{Reference}}.
    \underline{\cmdRefErrs[Hallucinated reference]}.
    \textcolor{gray}{The necessary context (e.g. known conditions, prior steps)}.
}
\label{tab:schema_examples}
\vspace{-16pt}
\end{table*}
\renewcommand{\arraystretch}{1} 

Second, proofs use references as justification for steps of reasoning; for instance, \hyperlink{https://proofwiki.org/wiki/Real_Addition_is_Commutative}{Real Addition is Commutative} provides justification for the statement $x+y=y+x$ where $x,y\in \mathbb{R}$, but not for $xy=yx$.
This aspect is analogous to using an article to  justify a claim (e.g. \citep{nakano2021WebGPT}).
Finally, proofs should not hallucinate references, or `beg the question' by self-referencing the current theorem.

\textbf{\cmdEqnErrs[Equations.]} 
Proofs contain a variety of multi-step derivations, ranging from  simple arithmetic to more sophisticated derivations (e.g. see \autoref{tab:err_eqn}).
A derivation should start with a valid equation given the surrounding context (e.g. $x+x=2x$ in \autoref{tab:schema_examples} versus $x+x=3x$).
Each subsequent step should be a valid derivation from the previous step, e.g. stating $=(2k+6)-1$ after $y=2k+5$.

\textbf{\cmdOtherErrs[Other reasoning], \cmdLangErrs[language], \& \cmdSymErrs[symbolic] errors.}
A proof should provide sufficient evidence that a claim is true to a human reader; it should not skip steps.
Proof steps should make progress towards proving the goal; in particular, they should not repeat known conditions in the theorem or conclusions made in a prior step. 
Finally, our schema leaves room for any other reasoning errors, as well as symbol errors (e.g. undefined symbols) and language errors (e.g. incomplete statements).

\textbf{\cmdOverallUseful[Usefulness] and \cmdOverallCorrect[correctness.]}
To judge the potential utility of language models as assistive systems in natural mathematics, we measure whether generated next-steps and full proofs are potentially useful hints for proving the theorem on one's own.
Additionally, we measure a summary judgment of correctness.
Note that an incorrect statement can still be helpful; for instance, it could give a hint for the type of reference to use, derivation to perform, argument to make, etc.

\textbf{Human evaluation protocol.} We measure these aspects through human annotation at a \textit{step-wise} and an \textit{overall} level.
For a step-wise annotation, an annotator is presented with the theorem, proof-so-far, and a generated next-step.
The annotator labels the $\{0,1\}$ correctness, usefulness, and presence of fine-grained errors outlined above.
After labeling each step of a proof, the annotator rates the full proof's overall correctness and usefulness on a 0-5 scale.
A rating of 4 or 5 is needed to be considered as correct, and a rating of 3 or above is needed to be considered as useful.

\textbf{Automatic metrics: lexical content.}
As automatic proxies for quality, we compare each generated proof against its ground-truth counterpart using the sentence-level $n$-gram matching metric \textsc{Gleu} \citep{mutton2007gleu}, and following work in knowledge-grounded dialogue \citep{shuster-etal-2021-retrieval-augmentation} we use F1 overlap between generated and ground-truth tokens.
Prior to computing the metrics, we normalize the generated and ground-truth proofs by only keeping the surface form of references, removing formatting characters with a MediaWiki parser, and collapsing any consecutive whitespace into a single space.

\textbf{Automatic metrics: knowledge grounding.}
We define knowledge grounding as meaning that a generated proof contains the same references as those found in the ground-truth proof.
To measure this, we use precision, recall, and F1-score between the reference sets contained in the generated and ground-truth proofs; i.e. $m(\{\hat \rb_1,\ldots,\hat \rb_{\hat R}\}, \{\rb_1^*,\ldots,\rb_{R_*}^*\})$, where $m(\cdot)$ is precision, recall, or F1.
We also use Knowledge Token-F1 (kF1) (\citep{shuster-etal-2021-retrieval-augmentation}), the overlap of the generated proof's tokens with tokens contained in the references mentioned in the ground-truth proof.

\section{Experiments}
\label{sec:experiments}

We use the training and dev splits of \textsc{NaturalProofs-Gen} during fine-tuning,
and the \textit{core evaluation sets} consisting of 100 theorems from the validation set and 100 from the test set for evaluation (see \S\ref{sec:task}).
These theorems were selected by the authors such that by looking at the theorem title each author could recall its content and sketch a proof.
While this may shift the evaluation towards an easier slice of the dataset, it was necessary to make human evaluation at a meaningful scale feasible. 
We also use the core sets for explorations and ablations.

We finetune three GPT-3~\citep{brown2020gpt3} (Curie) models,
using the OpenAI API (see \autoref{sec:implementation} for details):
\setdefaultleftmargin{1.5em}{}{}{}{}{}
\begin{enumerate}[topsep=1pt,itemsep=1pt,partopsep=1pt, parsep=1pt]
    \item \textbf{Baseline GPT-3.} We finetune a baseline GPT-3 model, $p_\theta(\yb|\xb)$, on theorem-proof examples $\{(\xb,\yb)\}$ from the training split. 
At test time, we condition the model on a test theorem.
    \item \textbf{$\methodnamenospace_{\textsc{Retrieve}}$.} 
    We finetune GPT-3 with retrieved references, $p_\theta(\yb|\xb,\hat{\rb}_1,\ldots,\hat{\rb}_{20})$.
We use a pretrained joint retrieval model $f(\xb)\rightarrow (\rb_1,\ldots,\rb_{|\mathcal{R}|})$ from \citep{welleck2021naturalproofs}, which was trained to retrieve an input theorem's  ground truth references. 
At test time, the model receives a  theorem and the top-20 reference titles that are retrieved given the theorem.
    \item \textbf{$\methodnamenospace$.} We finetune GPT-3 with human-provided references, $p_\theta(\yb|\xb,\rb_1^*,\ldots,\rb_{R_{\yb}}^*)$, where $\{\rb_1^*,\ldots,\rb_{R_{\yb}}^*\}$ is the set of reference-titles in the ground-truth proof.
    We use reference-title conditioned examples (Eqn.~\ref{eqn:ref-cond}) and reference-reconstruction (Eqn.~\ref{eqn:ref-rec}) on the training split/reference set.
    At test time, the model receives a theorem and reference titles from its ground-truth  proof.
\end{enumerate}
For \textbf{next-step suggestion} we use the human-provided knowledge model ($\methodnamenospace$).

\textbf{Decoding.} 
For full proof generation, we use stepwise++ decoding with the provided knowledge model, which we refer to as \textbf{$\methodname_{++}$}, and otherwise use greedy decoding.
We do not use stepwise constrained decoding with retrieved references since these references introduce noisy constraints, nor for next-step prediction since the algorithm is designed for multi-step proofs.
See \S\ref{sec:implementation} for additional experimental details.

\begin{table}[t]
\centering
\small
\setlength{\tabcolsep}{3pt}
\begin{tabular}{lrrrrrrrrr}
\toprule
& \multicolumn{3}{c}{Reasoning Errs $(\downarrow)$} & \multicolumn{2}{c}{Lexical Errs $(\downarrow)$} & \multicolumn{2}{c}{Per-Step $(\uparrow)$} & \multicolumn{2}{c}{Full Proof ($\uparrow$)} \\
& \cmdRefErrs[Ref.] & \cmdEqnErrs[Eqn.] & \cmdOtherErrs[Other] & \cmdLangErrs[Lang.] & \cmdSymErrs[Sym.] & \cmdStepUseful[Useful] & \cmdStepCorrect[Correct] & \cmdOverallUseful[Useful] & \cmdOverallCorrect[Correct] \\
\midrule
GPT-3    & 30.92 & 32.54 & 40.15 &  5.61 &  5.24 & 25.69 & 28.18 & 20\% & 13\% \\
$\methodnamenospace_{\textsc{Retrieve}}$ & \textbf{23.52} & 37.55 & 23.66 &  4.54 &  6.19 & 41.54 & 33.56 & 32\% & 24\% \\
\midrule
\methodnamenospace  & 25.84 & 35.93 & 25.23 &  8.41 &  5.35 & 39.60 & 26.30 & 35\% & 24\% \\
\mnamepp & 23.61 & \textbf{28.54} & \textbf{18.45} & 5.58 & 3.65 & \textbf{46.57} & \textbf{35.41} & \textbf{45\%} & \textbf{32\%} \\
\midrule
Next-step {\tiny (\methodnamenospace)}   & 19.70 & 26.32 & 19.10 &  8.57 &  5.86 & 51.43 & 42.86 & -- & -- \\
\bottomrule
\end{tabular}
\vspace{4pt}
\caption{Human evaluation results on the core test set for full proof generation and next-step suggestion (bottom row).
All models are fine-tuned on $\textsc{NaturalProofs-Gen}$. 
Knowledge -- either retrieved or human provided -- and constrained decoding improve proof generation, with 46\% of proof steps rated as useful and 35\% correct according to university-level mathematics students.
}
\label{tab:results_human_main}
\vspace{-24pt}
\end{table}
\textbf{Human evaluation setup.} To evaluate the proofs generated by \methodnamenospace, we recruited 15 students from the Department of Mathematics and Applied Mathematics at the University of 
Washington,
including undergraduate, masters, and Ph.D. students.
The annotators were trained on how to evaluate proof correctness and compensated according to IRB requirements;
see \S\ref{sec:additional_human}.
For each task, we first reveal the theorem and its gold proof to the annotator.
If they cannot understand a theorem or its gold proof, they may skip evaluating it.
Otherwise, they may proceed to see the model-generated proof, one step at a time, and annotate each step under the step-wise evaluation schema (outlined in \S\ref{sec:evaluation}).
After all the steps are shown and evaluated, for the full-proof generation task, the annotator is asked to annotate the entire proof under the overall evaluation schema.

\subsection{Main Results}
\label{sec:main_results}

\textbf{Our best method is capable of generating correct and useful proofs.}
According to human evaluation results (\autoref{tab:results_human_main}), our best method is \methodname with human-provided references and stepwise++ inference.
32\% of the proofs generated by this method are rated as \cmdOverallCorrect[correct] and 45\% are rated as \cmdOverallUseful[useful] as an aid for human proof writers.
On the per-step level, 35\% of the proof steps are \cmdOverallCorrect[correct] and 47\% are \cmdOverallUseful[useful].
Taking a more granular view, our best method makes significantly less \cmdRefErrs[reference], \cmdEqnErrs[equation], and \cmdOtherErrs[other reasoning errors] than other baselines.
It makes very few \cmdLangErrs[language] and \cmdSymErrs[symbolic errors], meaning it produces mostly complete, well-formatted mathematical statements, and defines and uses symbols accordingly.
It mostly avoids hallucinating references or creating circular proofs.
On the other hand, despite improving over the other methods, the model often struggles with correctly deploying and using references (23.6\% reference error rate), as well as symbolic computations (28.5\% equation error rate), especially multi-step derivations (21.9\% invalid).

\begin{table}[t]
\centering
\begin{minipage}[t]{.68\linewidth}
\scriptsize
\setlength{\tabcolsep}{2pt}
\begin{tabular}[t]{lp{6.0cm}c}
\toprule
\textbf{Theorem} & \textbf{Singleton Set is not Dense-in-itself} \\

& Let \(T = \left({S, \tau}\right)\) be a \underline{topological space}.

Let \(x \in S\).

Then the \underline{singleton set} \(\left\{{x}\right\}\) is not \underline{dense-in-itself}.\\
\midrule

GPT-3 & 

Let \(x \in S\). & \textcolor{green}{\cmark} \\

& From \textcolor{black}{\underline{\cmdRefErrs[Clopen Points in Topological Space]}} we have that 

\quad \(\left\{{x}\right\}\) is \underline{open} in \(T\).&\textcolor{red}{\xmark}\\

& From \textcolor{black}{\underline{\cmdRefErrs[Point is Open iff it is not Dense-in-itself]}}, it follows

 \quad that
 \(\left\{{x}\right\}\) is not \underline{dense-in-itself}.& \multirow{2}{*}{\textcolor{red}{\xmark}}\\
\midrule

$\methodnamenospace_{++}$ &

From \textcolor{blue}{\underline{Singleton Point is Isolated}}, \(\left\{{x}\right\}\) has an \underline{isolated point}.&\textcolor{green}{\cmark} \\

& Hence the result by definition of \textcolor{blue}{\underline{dense-in-itself}}.&\textcolor{green}{\cmark} \\
\bottomrule
\end{tabular}
\end{minipage}
\hfill
\begin{minipage}[t]{.31\linewidth}
\scriptsize
\setlength{\tabcolsep}{2pt}
\centering
\begin{tabular}[t]{|p{4cm}|}
\toprule
\textcolor{blue}{\underline{Singleton Point is Isolated}}

Let $T = \left({S, \tau}\right)$ be a \underline{topological space}.

Let $x \in S$.

Then $x$ is an \underline{isolated point} of 

\quad the \underline{singleton set} $\left\{{x}\right\}$, 

but not necessarily an \underline{isolated point} of $T$.\\
\bottomrule
\end{tabular}

\begin{tabular}[t]{|p{4cm}|}
\toprule
\textcolor{blue}{\underline{Dense-in-itself}}

Let $T = {S, \tau}$ be a \underline{topological space}.

Let $H \subseteq S$.

Then $H$ is \underline{dense-in-itself} iff it contains no \underline{isolated points}.
\\
\bottomrule
\end{tabular}
\end{minipage}
\vspace{2pt}
\caption{
GPT-3 \textcolor{black}{\underline{\cmdRefErrs[hallucinates references]}}, while the knowledge-grounded $\methodnamenospace_{++}$ with constrained decoding \textcolor{blue}{\underline{correctly uses references}}, resulting in a correct and useful proof.}
\label{tbl:example-dense}
\vspace{-2.0em}
\end{table}

\paragraph{What do the model's correct proofs look like?} We inspected the proofs labeled as correct and found three main categories:
(1) \textit{reference-assembly} proofs whose correctness is heavily determined by reference statements (e.g. \autoref{tbl:example}, \autoref{tab:qual_reference});
(2) \textit{template-adaptation} proofs in which the model adapts the structure and content of a training theorem's proof to prove the unseen evaluation theorem (e.g. \autoref{tab:qual_template_equations}, \autoref{tab:qual_template_cases});
(3) \textit{complex} proofs that are not fully determined by reference statements and differ significantly from training proofs (e.g. \autoref{fig:fig1}, \autoref{tbl:example-dense}).
In terms of techniques, our method demonstrates some ability to produce direct proofs (\autoref{tab:qual_complex_direct}), proofs by cases (\autoref{tab:qual_template_cases}), proofs by induction (\autoref{tab:qual_complex_induction}), utilize references (\autoref{tab:qual_reference}) and do symbolic computations (\autoref{tab:qual_template_equations}).

\paragraph{Vanilla fine-tuned GPT-3 struggles with proof generation.}
The vanilla fine-tuned GPT-3 model yielded fewer \cmdOverallUseful[useful] and \cmdOverallCorrect[correct] proofs, with more \cmdRefErrs[reference-based] and \cmdOtherErrs[other reasoning errors] than all three knowledge-grounded settings.
The model showed severe reference hallucination (18\%) and repetition (23\%).
It also makes significantly more reasoning errors related to reference usage.
Language and symbolic error rates roughly stay the same.
Overall, naively fine-tuning GPT-3 on theorem-proof examples alone is suboptimal for proof generation.

\paragraph{Human-provided knowledge improves proof generation.}
Grounding the generations with human-provided references significantly raises \cmdOverallCorrect[correctness] and \cmdOverallUseful[usefulness] of the proofs in both full-proof and per-step evaluation.
It most substantially reduces \cmdRefErrs[reference errors], especially invalid deployments and hallucinated references.
For example, \autoref{tbl:example-dense} shows the model grounding a proof with information from the theorem \hyperlink{https://proofwiki.org/wiki/Singleton_Point_is_Isolated}{Singleton Point is Isolated} and the definition of \hyperlink{https://proofwiki.org/wiki/Definition:Dense-in-itself}{Dense-in-itself}, in contrast to the vanilla GPT-3 model which hallucinates references.

\paragraph{Retrieved knowledge also improves proof generation.}
Retrieved knowledge also turns out to be very helpful, and even comparable to human-provided knowledge in some metrics.
Although the retrieval model is far from perfect, the proof generation model is capable of narrowing down the retrieved reference titles provided in its context, assembling proofs that are \cmdOverallUseful[useful] and \cmdOverallCorrect[correct] more often than the no-knowledge model.
Qualitatively, we found examples where grounding in retrieved references eliminates repetition, enables multi-step derivations justified by references (\autoref{tab:qual_template_equations}), and assembles references into a correct proof (\autoref{tab:qual_reference}).
This paves a promising path towards fully automated mathematical proof generation in natural mathematical language.
\begin{wraptable}{r}{0.45\textwidth}
\setlength{\tabcolsep}{3pt}
\centering
\small 
\begin{tabular}[t]{cccc}
\toprule
In-context & Stepwise++ & PPL ($\downarrow$) & Ref-F1 ($\uparrow$)\\
\midrule
\textcolor{red}{\xmark} & \textcolor{red}{\xmark}     & 1.0639 & 26.33 \\
\textcolor{red}{\xmark} & \textcolor{green}{\cmark}   & 1.0549 & 30.07 \\
\midrule
\textcolor{green}{\cmark} & \textcolor{red}{\xmark}   & 1.0644 & 89.43 \\
\textcolor{green}{\cmark} & \textcolor{green}{\cmark} & 1.0549 & 94.25 \\
\bottomrule
\end{tabular}
\caption{
    Stepwise++  approximates the constrained objective better than greedy.
}
\label{tab:ablation_constrained_objective}
%
\end{wraptable}

\textbf{Constrained decoding further improves proof generation.}
\autoref{tab:ablation_constrained_objective} confirms that stepwise++ decoding approximates the constrained objective (Eqn.~\ref{eqn:constrained}) better than greedy search, yielding
proofs with lower perplexity and higher constraint satisfaction (Ref-F1). %
This translates to generations that are correct and useful more often according to the annotators.
Intuitively, the constraints encourage the model to include references that help prove the claim (e.g. \autoref{tbl:example}).

\paragraph{Next-step suggestion.}
The next-step suggestion task characterizes a model's performance on making a single proof step given a correct proof-so-far.
In \autoref{tab:results_human_main} we use the provided-knowledge model with greedy decoding for next-step suggestion, and find that reasoning errors decrease and per-step usefulness and correctness improve compared to the full proof setting, with 51\% of the proof steps rated as useful and 43\% correct.
Although we used a single suggestion in our human evaluation study, in \autoref{tab:results_multi_main} we simulate a user choosing from among multiple suggestions by sampling 10 next-steps from our model and 
computing automatic metrics on the sample with the best sum of metrics.
Using 10 samples instead of greedily decoding a single sequence substantially improves each metric, suggesting that utility might be increased further by presenting  multiple suggestions.

\begin{wraptable}{r}{0.34\textwidth}
\vspace{-1em}
\centering
\footnotesize
\setlength{\tabcolsep}{3pt}
\begin{tabular}[t]{lccccccc}
\toprule
\textbf{Decoding} & GLEU & Ref-F1 \\
\midrule
Greedy & 47.87 &  65.50  \\
\midrule
Temp (t=.6) & 60.60  &84.44  \\
Temp (t=.8) & 61.89  & 86.74  \\
Temp (t=1.0) & \textbf{62.12} & \textbf{86.87}  \\
\bottomrule
\end{tabular}
\vspace{2pt}
\caption{
    \textit{Next-step suggestion}: Sampling 10 suggestions improves over a single greedy suggestion.
}
\label{tab:results_multi_main}
\vspace{-1em}
\end{wraptable}

\paragraph{How good are Automatic Metrics?}
\label{sec:auto_metrics_results}
\begin{table}
\centering
\footnotesize
\setlength{\tabcolsep}{4pt}
\begin{tabular}[t]{lccccccc}
\toprule
& \multicolumn{2}{c}{Lexical} & \multicolumn{5}{c}{Grounding} \\
\cmidrule(lr){2-3}\cmidrule(lr){4-8}
& GLEU & Token F1 & kF1 & Ref-P & Ref-R & Ref-F1 & Halluc ($\downarrow$) \\
\midrule
GPT-3 & 24.40 & 49.96 & 49.30 & 29.93 & 24.73 & 23.69 & 17.92 \\
$\methodnamenospace_{\textsc{Retrieve}}$ & 26.58 & 53.02 & 55.88 & 38.17 & 28.48 & 27.10 & 2.25 \\
\methodnamenospace & 35.27 & 66.00 & 90.07 & 93.05 & 86.05 & 87.08 & 1.60 \\
\mnamepp & 34.49 & 65.61 & 96.39 & 94.66 & 95.00 & 93.92 & 1.71 \\
\midrule
\includegraphics[scale=0.25]{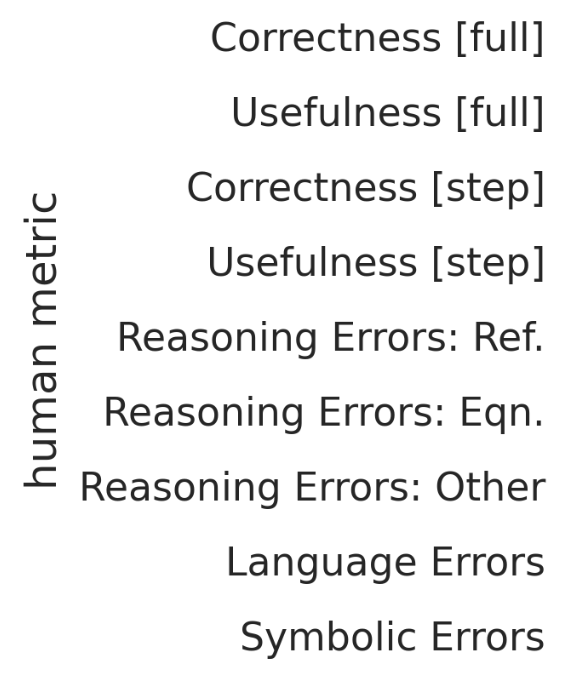} & \includegraphics[scale=0.25]{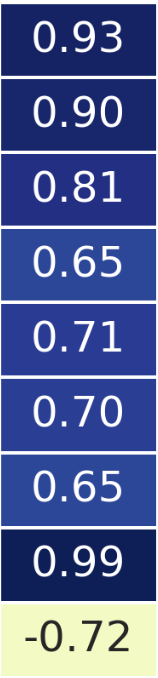} & \includegraphics[scale=0.25]{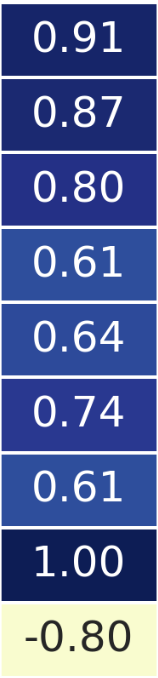} & \includegraphics[scale=0.25]{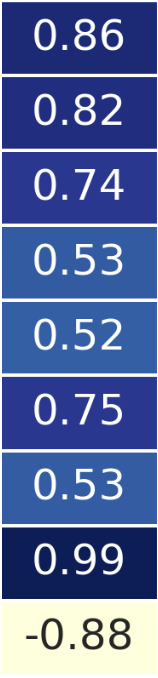} & \includegraphics[scale=0.25]{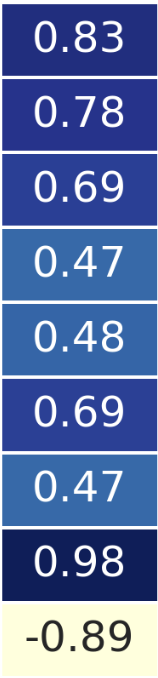} & \includegraphics[scale=0.25]{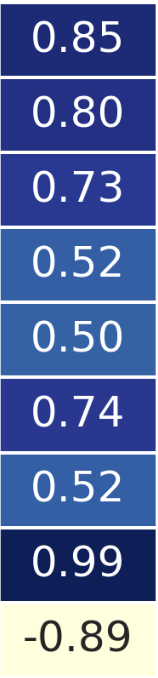} & \includegraphics[scale=0.25]{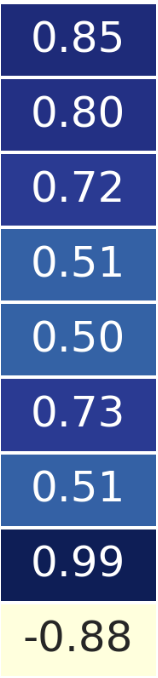} & \includegraphics[scale=0.25]{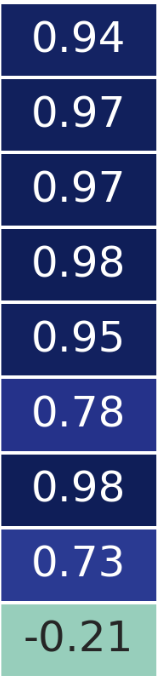} \\
\bottomrule
\end{tabular}
\vspace{2pt}
\caption{
    Automatic metrics on the core test set for full-proof generation, and correlation between human metrics and automatic metrics on the core validation set.
}
\label{tab:results_auto_main}
\vspace{-16pt}
\end{table}
We study how well the automatic lexical and grounding metrics introduced in (\S\ref{sec:evaluation}) can reflect the real quality of proofs, as a guide for using them as a proxy evaluation protocol for \textsc{NaturalProofs-Gen}.
We compute the Pearson correlation coefficient between each pair of human and automatic metrics, with data from the four experiment settings for full-proof generation.
Results are shown in the lower part of \autoref{tab:results_auto_main}, with error metrics negated, meaning positive correlation is desired.

The lexical and grounding metrics positively correlate with full proof \cmdOverallCorrect[correctness] and \cmdOverallUseful[usefulness] ($\geq$ 0.8).
At the step-level, the metrics show (i) high correlation with step-level \cmdOverallCorrect[correctness] and \cmdLangErrs[language errors]; (ii) varied, but positive, correlations with aggregate reasoning errors; (iii) negative correlation with \cmdSymErrs[symbolic errors] (though symbolic errors are relatively low for all models).
The results suggest that optimizing for automatic metrics may be a viable strategy,
albeit without guarantees on how finer-grained reasoning aspects vary across proofs.

\subsection{Ablations and error analysis.}
\label{sec:further_analysis}

\begin{wraptable}{r}{0.37\textwidth}
\vspace{-1em}
\setlength{\tabcolsep}{3pt}
\begin{center}
\begin{tabular}[t]{c|ccc}
\toprule
Recon. & Gleu & Ref-F1 & Halluc.\\
\midrule
\xmark & 33.03 & 82.85 & 3.32\\
\cmark & \textbf{35.93} & \textbf{84.15} & \textbf{2.68}\\
\bottomrule
\end{tabular}
\caption{Effect of reference reconstruction in \methodname 
(greedy decoding, full validation set).}
\label{tbl:recon}
\end{center}
\vspace{-1em}
\end{wraptable}

\textbf{Reference reconstruction.}
We fine-tune an additional GPT-3 model that is provided with in-context reference titles, but without reference reconstruction.
As seen in \autoref{tbl:recon}, 
reference reconstruction improves content and reference usage.

\textbf{Constrained decoding.}
First, \autoref{tbl:stepwise-segment} compares the step-level search in stepwise++ with searching at the full-proof level through sampling multiple proofs and selecting the best with the \methodname value function (\textit{rerank (n)}).
Reranking 60 samples matches the cost of stepwise++ in terms of number of decoded tokens.
Full-proof reranking yields the best Gleu, though with lower reference-F1.
Second, \autoref{tbl:stepwise-mechanism} shows that the expansion and selection mechanisms together result in the best reference matching, while holding Gleu at a similar level.
\begin{table}[t]
\begin{minipage}[t]{.48\linewidth}
\setlength{\tabcolsep}{3pt}
\begin{center}
\begin{tabular}[t]{ccll}
\toprule 
Expand & Select & GLEU & Ref-F1\\
\midrule
\xmark & \xmark  & 40.62 (.84)&91.78 (.49)\\
\cmark & \xmark  & 41.12 (.58)&92.61 (.63)\\
\xmark & \cmark  & 39.14 (.55)&93.11 (.34)\\
\cmark & \cmark  & 40.11 (1.55) &\textbf{94.13} (.45)\\
\bottomrule
\end{tabular}
\end{center}
\caption{Ablation of the stepwise++ expansion and selection mechanisms. Mean (std) over 3 runs shown on the core dev set.}
\label{tbl:stepwise-mechanism}
\end{minipage}
\hfill
\begin{minipage}[t]{.50\linewidth}
\setlength{\tabcolsep}{3pt}
\begin{center}
\begin{tabular}[t]{cll}
\toprule
Decoding  & Gleu & Ref-F1\\
\midrule
Greedy & 41.12 (--)& 89.30 (--) \\
Rerank (10) & \textbf{43.88} (.29) & 91.72 (.28)\\
Rerank (60) & 42.23 (.80) &93.16 (.27)\\
Stepwise++  & 40.11 (1.55) &\textbf{94.13} (.45)\\
\bottomrule
\end{tabular}
\end{center}
\caption{Stepwise versus full-proof search. 
Mean (std) over 3 runs on the core dev set.}
\label{tbl:stepwise-segment}
\end{minipage}
\vspace{-24pt}
\end{table}

Finally, \autoref{tbl:value-alpha} shows that both terms in the \methodname value function $\alpha v_{\text{constraints}} + (1-\alpha)v_{\text{LM}}$ are needed: increasing the constraint weight $\alpha$ increases reference-matching, with a tradeoff in Gleu at high values.

\paragraph{Language model comparison.}
\autoref{tbl:lm-ablation} varies the language model used to parameterize \methodname.
The content and reference usage metrics improve with larger models.
Separately, 
we find that increasing inference-time compute closes the gap in reference-matching between GPT-2 and the larger GPT-3 model (\autoref{tbl:inference-budget}): sampling 10 full-proofs from GPT-2 and selecting the best using the \methodname value function achieves the same reference-F1 as GPT-3 with a single greedily-decoded proof.
However, Gleu remains much higher with the larger GPT-3 model.

\paragraph{Challenge: Reasoning with references.}
Although reference reasoning errors were decreased through knowledge-grounding and constrained decoding, \methodname still commits a reference error on 23.6\% of test steps (27\% dev), with 15\% of steps containing invalid deployments and 10\% invalid justifications. 
For next-step prediction, the reference error rate remains nontrivial (19.7\% test, 13\% dev).
, meaning that the model can struggle to correctly deploy references or use them as justification even in the absence of compounding errors from previous steps.
\autoref{tab:err_ref} shows example invalid deployments and justifications; the errors are at times subtle, and require reasoning about the theorem statement, reference content, and proof context.

\paragraph{Challenge: Equations and derivations.} 
\methodname commits an equation-related error on 28.5\% of test steps (22.8\% dev), including invalid equations  (9.4\%) and derivations (21.9\%). 
Though an improvement over vanilla fine-tuned GPT-3 (32.5\%), the errors occur frequently and remain high for next-step prediction (26\%).
\autoref{tab:err_eqn} shows representative errors, which range from simple `commonsense' mistakes (e.g. $24=2^3)$ to making invalid steps with false justification within more sophisticated multi-step proofs.
Investigating the role of pretraining, in-context techniques~\citep{nye2021ShowYW}, and autoformalization~\citep{szegedy2020promising} is interesting future work.
\begin{wrapfigure}{r}{0.45\textwidth}
\vspace{-0.5em}
\centering
\includegraphics[width=0.35\textwidth]{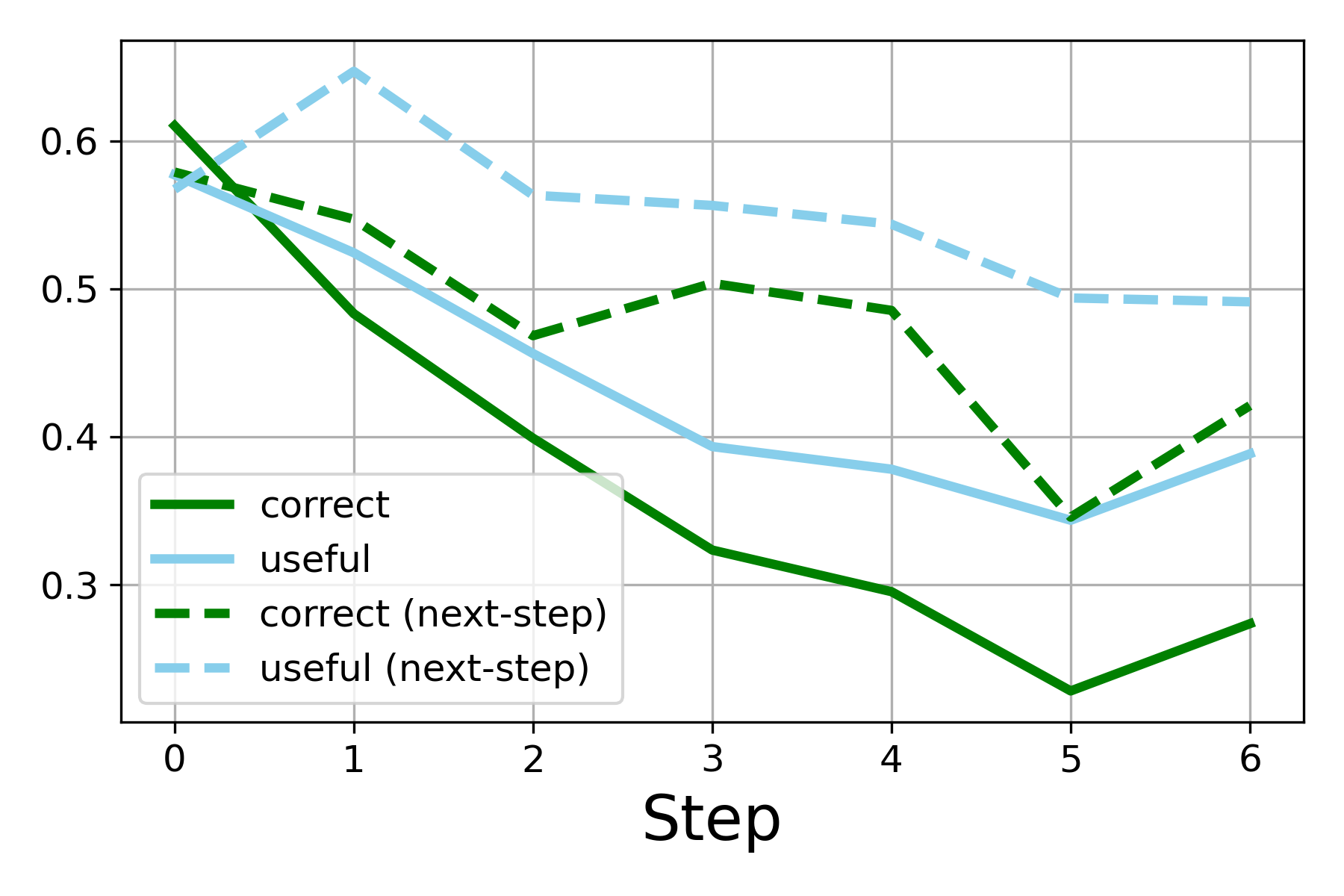}
\caption{Per-step correctness and usefulness as a function of step number, for full-proof generation with \mnamepp and next-step prediction with \methodnamenospace.}
\label{fig:results_perstep}
\vspace{-2em}
\end{wrapfigure}

\textbf{Challenge: Proof length.} Although \methodname demonstrates some ability to write long proofs (e.g. \autoref{tab:qual_complex_induction}), the 42\% next-step correctness suggests that compounding errors are likely as proof length increases.
Indeed, our best model's full-proof correctness is 48\% on 1-4 step proofs ($n=102$), decreasing to 15.6\% on proofs with 5 or more steps ($n=64$), with lower per-step usefulness and correctness at later steps (\autoref{fig:results_perstep}).
Our findings are analogous to recent work on language modeling for formal theorem proving \citep{polu2022formal}, where current models  are typically limited to chaining 2 or 3 non-trivial steps of mathematical reasoning.

\subsection{Additional discussion}
Finally, we provide higher-level comments on future work related to interactive systems, mathematical assistants, and generating proofs in informal versus formal mathematics.

\paragraph{Interactive \& improving systems.} Currently, our tasks are at two ends of a spectrum: in next-step generation, we always assume previous steps are from a human-written proof, while in full proof generation they are always from the model. 
Our results with multiple next-step suggestions suggest that users might find \textit{some} suggestion among the multiple returned useful at a high rate, pointing to a middle ground:
a human-in-the-loop \methodnamenospace, in which a human picks the next step from among the returned suggestions, or writes one based on the suggestions.
The selected or written next-step could then be used as feedback to improve the system, enabling an iteratively improving \methodnamenospace.
This notion of a continuously improving, teachable system is an emerging (e.g. \cite{dalvi2022TowardsTR}) and interesting future direction. 

\paragraph{Assistants for mathematics.}  Our tasks were motivated by an assistant that helps a user write a proof, either from scratch or when stuck part of the way through.
Our study here focuses on \textit{capability}: investigating whether neural language models are capable of performing the underlying mathematics that would be expected from such an assistant.
A further challenge is to also ensure \textit{reliability} -- a user should have confidence that the model is not deceptive or incorrect, and is robust to changes in domain, on nearby problems, and on alternative ways of expressing a problem.
Even further, we would like \textit{flexibility} -- human teachers can interact with a student flexibly through dialogue, natural language, and diagrams, rather than the strict input-output format defined by a dataset.
Our work provides an initial step towards this larger vision.

\paragraph{Informal and formalized mathematics.} Our work investigates theorem proving entirely in natural mathematical language (i.e. `informal' mathematics), as it reflects an interface that a student typically uses when working with mathematics.
An alternative is proving theorems in a formalized system, in which proof steps are expressed in a programming language (e.g. Lean~\cite{demoura2015lean}). 
Operating purely in a formalized system allows for verifying correctness -- unlike our setting which must be verified by a human -- arguably at the cost of flexibility and interpretability, as the mathematics is no longer expressed in natural language and must adhere to constraints of the formal system.
Investigating combinations of the two -- e.g. expressing a theorem in natural language, receiving a verified formal proof, then providing an interpretation in natural language -- presents a wide range of interesting directions for future work. 

\section{Related Work}
\paragraph{Formalized mathematics with neural language models.}
A large portion of work on  machine learning for mathematics focuses on  formalized mathematics.
Language models have been used for interactive theorem proving, including in GPT-$f$ \citep{polu2020generative, polu2022formal}, PACT \citep{han2021proof}, and in \cite{urban2020first}.
In these settings proof steps are expressed in a programming language (e.g. Lean \citep{demoura2015lean}) and there is access to a verifier, which differs from our setting of theorem proving in natural mathematical language.

\paragraph{Informal mathematics with neural language models.}
Previous work on theorem proving in natural mathematical language focuses on retrieving relevant premises (e.g. theorems, definitions) \citep{ferreira2020natural,ferreira2020premise,welleck2021naturalproofs,han2021contrastive}, or informal-to-formal translation \citep{wang2020exploration}, which differ from our setting of generating next-steps or full proofs.
Outside of theorem proving, various works use sequence models for problem solving, including benchmarking language models on arithmetic \citep{saxton2018analysing} or competition problems \citep{hendrycks2021measuring},   symbolic mathematics~\citep{lample2020deep,welleck2022SymbolicBI}, augmenting LMs with verifiers~\citep{cobbe2021gsm8k} or in-context rationales~\citep{wei2022chain} for math word problems, or using language models for math-related program synthesis  \citep{austin2021program, drori2022neural} and competitive programming \citep{li2022competition}.
These settings focus on generating executable programs or a numerical answer, which differ from our theorem proving setting, where the goal is to generate sound and convincing arguments on a range of topics in natural mathematical language.

\paragraph{Related areas in NLP.}
Systematic reasoning in natural language (outside of math) has been studied with synthetic proofs~\citep{saha2020prover,tafjord-etal-2021-proofwriter}, single-step deductions~\citep{bostrom2021flexible}, or entailment trees~\citep{dalvi-etal-2021-explaining}, which differ from proving real-world mathematical theorems. 
Augmenting LMs with knowledge reduces hallucinations in dialogue~\citep{shuster-etal-2021-retrieval-augmentation} which has an analogous step-wise structure, while \citep{nakano2021WebGPT} use references within long-form answers; these and related NLP findings differ from improving the utility of mathematical proofs.
Lexically-constrained decoding algorithms include variants of (token-level) beam search (e.g. \citep{anderson-etal-2017-guided,hokamp-liu-2017-lexically,lu2021neurologic,lu2021NeuroLogicAD}) which assume access to per-token logits, and gradient-based decoding \citep{qin2022COLDDE}; our segment-level decoding only assumes a  sampler that returns text and its log-probability, making it compatible with recent language model API interfaces (e.g. the GPT-3 API).
\section{Conclusion}

We described \methodnamenospace, a knowledge-grounded language model that generates mathematical proofs by conditioning on background theorems and definitions, and optionally enforces their presence with constrained decoding.
Our system improves the quality of next-step suggestions and generated proofs over fine-tuned GPT-3, demonstrating an ability to correctly prove theorems and provide useful suggestions to human proof writers.

\begin{ack}
This work was funded in part by the Natural Sciences and Engineering Research Council of Canada (NSERC) (funding reference number 401233309), DARPA MCS program through NIWC Pacific (N66001-19-2-4031), and the Allen Institute for AI. We also thank Google Cloud Compute, as well as OpenAI.

The authors would like to thank Alisa Liu, Julian Michael, Yuren (Rock) Pang, and Kaiming Cheng for dogfooding and providing valuable feedback to our human evaluation system.
We would also like to thank James McGivern for developing an interactive demo for NaturalProver.
\end{ack}

\bibliographystyle{abbrvnat}
\bibliography{neurips_2022}

\newpage
\appendix

\section{Additional Results}
\label{sec:additional_results}

\subsection{Additional ablations}
\label{sec:additional_ablations}
\autoref{tbl:lm-ablation} shows automatic metrics with various language models used to parameterize \methodnamenospace.

\autoref{tbl:inference-budget} shows results with the 774M parameter GPT-2 model with greedy decoding, and full-proof sampling \& reranking with 5 and 10 samples, compared to the 13B parameter GPT-3 with greedy decoding.
We use $\tau=0.3$ and $\alpha=0.75$ based on our full-proof sampling experiments with GPT-3.

\autoref{tbl:value-alpha} varies the value function parameter $\alpha$ (core dev set).
We use full-proof sampling since stepwise++ uses multiple values of $\alpha$ in its selection.

\begin{table}[h]
\begin{minipage}[t]{.48\linewidth}
\setlength{\tabcolsep}{3pt}
\begin{center}
\begin{tabular}[t]{ ll|ccc }
  \toprule
  \textbf{Model} & \textbf{Params} & \textbf{Gleu} & \textbf{Ref-F1} & \textbf{Halluc} \\\toprule
  GPT-Neo & 125M & 24.85 & 61.42 & 11.07\\
  GPT-2 & 774M & 32.06 & 65.22 & 6.76\\
  GPT-J & 6B  & 39.14 & 79.23 & 3.51\\
  GPT-3 & 13B & \textbf{42.39} & \textbf{89.29} & \textbf{1.90}\\
  \bottomrule
\end{tabular}
\end{center}
\caption{Varying the language model parameterization of \methodname (provided knowledge, greedy decoding, core dev set).}
\label{tbl:lm-ablation}
\end{minipage}
\hfill
\begin{minipage}[t]{.50\linewidth}
\setlength{\tabcolsep}{3pt}
\begin{center}
\begin{tabular}[t]{ ll|ccc }
  \toprule
  \textbf{Model} & \textbf{Decoding} & \textbf{Gleu} & \textbf{Ref-F1} & \textbf{Halluc} \\\toprule
  GPT-2 & Greedy & 32.06 & 65.22 & 6.76\\
  GPT-2 & Rerank (5)  & 32.95 & 83.55 & 5.24\\
  GPT-2 & Rerank (10) & 32.65 & \textbf{89.30} & 2.89\\
  GPT-3 & Greedy & \textbf{42.39} & \textbf{89.29} & \textbf{1.90}\\
  \bottomrule
\end{tabular}
\end{center}
\caption{Increasing the inference-time compute budget and reranking with the \methodname value function closes the reference-matching gap between GPT-2 (774M) and GPT-3 (13B).
}
\label{tbl:inference-budget}
\end{minipage}
\end{table}

\begin{table}[h]
\setlength{\tabcolsep}{3pt}
\begin{center}
\begin{tabular}[t]{ccc}
\toprule
$\alpha$ & \textbf{Gleu} &\textbf{Ref-F1}\\
\midrule
0.0 & 42.79 & 88.40 \\
.25 & 42.05 & 90.81 \\
.50 & 42.59 & 91.75 \\
.75 & 42.17 & 93.19 \\
1.0 & 41.90 & 93.60 \\
\bottomrule
\end{tabular}
\end{center}
\caption{Effect of value function, from $\alpha:0$ (LM only) to $\alpha:1.0$ (constraint only),
with full-proof sampling (10).
}
\label{tbl:value-alpha}
\end{table}

\begin{table}[h]
\begin{center}
\footnotesize
\setlength{\tabcolsep}{4pt}
\begin{tabular}[t]{lccccccc}
\toprule
& \multicolumn{2}{c}{Lexical} & \multicolumn{5}{c}{Grounding} \\
\cmidrule(lr){2-3}\cmidrule(lr){4-8}
& GLEU & Token F1 & kF1 & Ref-P & Ref-R & Ref-F1 & Halluc ($\downarrow$) \\
\midrule
Stepwise Stochastic Beam  & 41.0 & 68.89 & 90.33 & 91.43 & 82.04 & 84.21 & 4.60 \\
Constrained Stepwise++ & 40.4 & 68.90 & 97.24 & 95.05 & 94.85 & 94.15 & 2.00\\
\bottomrule
\end{tabular}
\end{center}
\caption{NaturalProver with a stepwise stochastic beam search baseline versus stepwise++ decoding. The baseline search corresponds to using stepwise decoding with an LM-only value function ($\alpha:0$). 
Constrained stepwise++ decoding substantially improves grounding metrics compared to stochastic beam search, while keeping the lexical content metrics at a similar level.
Core validation set.
}
\label{tbl:value-alpha2}
\end{table}

\subsection{Multiple next-step suggestions}
\begin{table}
\centering
\footnotesize
\setlength{\tabcolsep}{4pt}
\begin{tabular}[t]{lccccccc}
\toprule
& \multicolumn{2}{c}{Lexical} & \multicolumn{5}{c}{Grounding} \\
\cmidrule(lr){2-3}\cmidrule(lr){4-8}
\textbf{Decoding} & GLEU & Token F1 & kF1 & Ref-P & Ref-R & Ref-F1 & Halluc ($\downarrow$) \\
\midrule
Greedy & 47.87 & 65.33 & 70.03 & 80.04 & 72.78 & 65.50 & 0.93 \\
\midrule
Nucleus (p=.5) & 51.10 & 68.34 & 73.69 & 82.75 & 74.93 & 69.21 & 0.94 \\
Nucleus (p=.7) & 53.97 & 71.01 & 78.86 & 84.75 & 79.28 & 74.52 & 0.66 \\
Nucleus (p=.9) & 57.79 & 74.45 & 85.66 & 90.17 & 84.03 & 81.83 & \textbf{0.22} \\
Temperature (t=.6) & 60.60 & 76.43 & 87.46 & 91.03 & 87.48 & 84.44 & 0.62 \\
Temperature (t=.8) & 61.89 & 77.48 & 89.67 & \textbf{93.19} & 88.46 & 86.74 & 0.43 \\
Temperature (t=1.0) & \textbf{62.12} & \textbf{77.60} & \textbf{89.78} & 93.05 & \textbf{88.96} & \textbf{86.87} & 0.63 \\
\bottomrule
\end{tabular}
\vspace{2pt}
\caption{
    Automatic metrics on the core dev set for next-step suggestion, with different decoding strategies.
    Generations are from $\methodnamenospace$ (provided knowledge).
    For stochastic decoding, 10 candidates are sampled. 
    We compute metrics using the candidate with the highest sum of all metrics (with Hallucination negated).
}
\label{tab:results_multi}
\end{table}
\autoref{tab:results_multi} shows next-step suggestion results with 10 sampled suggestions versus greedy decoding.

\newpage

\section{Qualitative Examples}
\label{sec:qual}
\subsection{Error Analysis}
\label{sec:errors}

\subsubsection{Reference errors}

{\scriptsize
\centering
\setlength{\tabcolsep}{2pt}
\begin{tabular}[t]{lp{9.0cm}cccccccccc}
\toprule
\textbf{Theorem 1} & \textbf{Geometric Mean of two Positive Real Numbers is Between them} \\
& Let $a, b \in \mathbb{R}$ be \underline{real numbers} such that $0 < a < b$.\\
&Let $ G ({a, b})$ denote the \underline{geometric mean} of $a$ and $b$.\\
&Then: 
 
 \quad $a <  G ({a, b}) < b$ \\
\midrule
\mnamepp & \textbf{Proof:} & correct& deploy \\
 & By definition of \textcolor{blue}{\underline{geometric mean}}: 
 
 \quad\cmdRefErrs[$ G ({a, b}) = \dfrac {a^2 + b^2} 2$] & \textcolor{red}{\xmark}& \textcolor{red}{\xmark}\\
 &...\\\cdashline{1-5}[0.2pt/4pt]\\
\textit{Gold Proof} & \textbf{Proof:} &  \\
& By definition of \textcolor{blue}{\underline{geometric mean}}: 
 
 \quad \cmdStepCorrect[$ G ({a, b}) := \sqrt {a b}$] 
 
 where $\sqrt {a b}$ specifically denotes the \underline{positive square root} of $a$ and $b$.\\
 & ... \\\cdashline{1-5}[0.2pt/4pt]\\
\textit{Reference} & \textcolor{blue}{{\underline{Geometric Mean}}}\\
&Let $x_1, x_2, \ldots, x_n \in \mathbb{R}_{>0}$ be \underline{(strictly) positive real numbers}.\\
&The geometric mean of $x_1, x_2, \ldots, x_n$ is defined as:\\

& \quad $G_n := ({\prod_{k = 1}^n x_k})^{1/n}.$\\
\end{tabular}
\begin{tabular}[t]{lp{9.0cm}cccccccccc}
\toprule
\textbf{Theorem 2} & \textbf{Power Set of Transitive Set is Transitive} \\

& Let $S$ be a \underline{transitive set}.\\
&Then its \underline{power set} $\mathcal{P}(S)$ is also a \underline{transitive set}. \\
\midrule
\mnamepp & \textbf{Proof:} & correct& deploy \\
& From \underline{Set is Transitive iff Subset of Power Set}: 
 
 \quad $S \subseteq \mathcal{P}(S)$ & \cmark& & \\
&From \textcolor{blue}{\underline{Power Set of Subset}}: 
 
 \quad \cmdRefErrs[$\mathcal{P}(S) \subseteq S$] & \textcolor{red}{\xmark}&  \textcolor{red}{\xmark}& \\
 &...\\\cdashline{1-5}[0.2pt/4pt]\\
\textit{Gold Proof} & \textbf{Proof:} &  \\
 & Let $S$ be \underline{transitive}.\\
&By \underline{Set is Transitive iff Subset of Power Set}: 
 
 \quad $S \subseteq \mathcal{P}(S)$\\
&Then by \textcolor{blue}{\underline{Power Set of Subset}}: 
 
 \quad \cmdStepCorrect[$\mathcal{P}(S) \subseteq \mathcal{P}(\mathcal{P}(S))$]\\
& ...\\\cdashline{1-5}[0.2pt/4pt]\\
\textit{Reference} & {\textcolor{blue}{\underline{Power Set of Subset}}}\\
& Let $S \subseteq T$ where $S$ and $T$ are both \underline{sets}.

Then:

\quad $\mathcal{P}(S) \subseteq \mathcal{P}(T)$
where $\mathcal{P}(S)$ denotes the \underline{power set} of $S$.\\
\bottomrule
\end{tabular}
\vspace{2pt}
\captionof{table}{\textit{Reference deployment errors.} Theorem 1: \methodname incorrectly deploys the definition of geometric mean  ($G(a,b)=(ab)^{1/2}$ would be correct).
Theorem 2: An invalid deployment of \texttt{Power Set of Subset}; a correct application would yield  $\mathcal{P}(S)\subseteq \mathcal{P}(\mathcal{P}(S))$.
All generated proofs are with provided knowledge and stepwise++ decoding.
}
\label{tab:err_ref}
}
{
\scriptsize
\setlength{\tabcolsep}{2pt}
\centering
\begin{tabular}[t]{lp{8.5cm}cccccccccc}
\toprule
\textbf{Theorem 3} & \textbf{Derivative at Point of Inflection} \\

& Let $f$ be a \underline{real function} which is \underline{twice} \underline{differentiable} on the \underline{open interval} $\left({a \,.\,.\, b}\right)$.\\

&Let $f$ have a \underline{point of inflection} at $\xi \in \left({a \,.\,.\, b}\right)$.\\

&Then: 
 
 \quad $f'' \left({\xi}\right) = 0$ 
 
 where $f'' \left({\xi}\right)$ denotes the \underline{second derivative} of $f$ at $\xi$. \\
\midrule
\mnamepp & \textbf{Proof:} & correct& deploy& justif. \\
 & By definition of \textcolor{blue}{\underline{point of inflection}}:

 \quad \cmdRefErrs[$\forall \eta \in \left({a \,.\,.\, b}\right): f \left({\eta}\right) = 0$]
 
 where $f \left({\eta}\right)$ denotes the \underline{derivative} of $f$ with respect to $\eta$. & \textcolor{red}{\xmark}& \textcolor{red}{\xmark}& \textcolor{red}{\xmark}\\
 & ...\\\cdashline{1-5}[0.2pt/4pt]\\
\textit{Gold Proof} & \textbf{Proof:} &  \\
 & By definition of \textcolor{blue}{\underline{point of inflection}},
 
 \quad \cmdStepCorrect[$f'$ has either a]\underline{\cmdStepCorrect[local maximum]}\cmdStepCorrect[or a]\underline{\cmdStepCorrect[local minimum]}\cmdStepCorrect[at $\xi$.]\\
&... \\\cdashline{1-5}[0.2pt/4pt]\\
\textit{Reference} & {\textcolor{blue}{\underline{Point of Inflection}}}\\
& Let $f$ be a \underline{real function} which is \underline{differentiable} on an \underline{interval} $\mathbb{I} \subseteq \mathbb{R}$.

Let $\xi \in \mathbb{I}$.

$f$ has a point of inflection at $\xi$ {{iff}} the \underline{derivative} $f'$ of $f$ 

\quad has either a \underline{local maximum} or a \underline{local minimum} at $\xi$.
\\
\end{tabular}
\begin{tabular}[t]{lp{8.5cm}cccccccccc}
\toprule
\textbf{Theorem 4} & \textbf{Minus One is Less than Zero} \\

& \quad$-1 < 0$ \\
\midrule
\mnamepp & \textbf{Proof:} & correct& useful& justif. \\
 &  
 $0 < 1 
        \quad\quad\quad\quad\ \underline{\text{Real Zero is Less than Real One}} 
$ & \cmark& \cmark& \\
&\cmdRefErrs[$\leadsto -1 <0$]$ 
        \quad\ \ \underline{\text{\textcolor{blue}{\cmdRefErrs[Order of Real Numbers is Dual of Order of their Negatives]}}} 
$ & \textcolor{red}{\xmark}& \cmark& \textcolor{red}{\xmark}\\
&$ \leadsto -1 < 0 
        \quad\quad\underline{\text{Negative of Real Zero equals Zero}}
$ 
  & \cmark& \cmark& \\
&$\blacksquare$ & &  \\\cdashline{1-5}[0.2pt/4pt]\\
\textit{Gold Proof} & \textbf{Proof:} &  \\
 &  
 $0 < 1 \quad\quad\quad\quad\ \ \underline{\text{Real Zero is Less than Real One}}$\\
&\cmdStepCorrect[$\leadsto -0  >  -1$]$ 
        \ \ \underline{\text{\textcolor{blue}{\cmdStepCorrect[Order of Real Numbers is Dual of Order of their Negatives]}}}$\\
&$\leadsto 0  >  -1 
        \quad\quad\underline{\text{Negative of Real Zero equals Zero}} 
$\\
&$\leadsto -1 < 0 
        \quad\quad\underline{\text{Dual Ordering}} $\\
&$\blacksquare$ \\\cdashline{1-5}[0.2pt/4pt]\\
\textit{Reference} & {\textcolor{blue}{\underline{Order of Real Numbers is Dual of Order of their Negatives}}}\\
& \quad $\forall x, y \in \mathbb{R}: x > y \iff ({-x})< ({-y})$\\
\bottomrule
\end{tabular}
\vspace{2pt}
\captionof{table}{\textit{Reference justification errors.} 
Theorem 3: \methodname makes an invalid inference using the definition of point of inflection (given the theorem statement, $f(\xi)=0$ holds, but not necessarily $f(\eta)=0$ for \textit{all} $\eta$ in the interval), which can be viewed as both incorrectly deploying the definition and using it as invalid justification.
Theorem 4: A subtle invalid justification that is still useful to the human annotator: the reference provides justification for $-0>-1$ rather than $-1<0$.
See \autoref{tab:err_eqn} for additional justification errors within derivations.
All generated proofs are with provided knowledge and stepwise++ decoding.
}
\label{tab:err_ref2}
}

\subsubsection{Equation errors}
{
\scriptsize
\setlength{\tabcolsep}{2pt}
\centering
\begin{tabular}[t]{lp{9.1cm}cccccccccc}
\toprule
\textbf{Theorem 1} & \textbf{Sum of Squares of Divisors of 24 and 26 are Equal} \\

& The \underline{sum} of the \underline{squares} of the \underline{divisors} of $24$ equals the \underline{sum} of the \underline{squares} of the \underline{divisors} of $26$: 

 $\quad {\sigma_2} ({24}) =  {\sigma_2} ({26})$ 
 
 where $\sigma_\alpha$ denotes the \underline{divisor function}. \\
\midrule
& \textbf{Proof:} & correct& \cmdEqnErrs[eqn.]\\
\mnamepp &  
\cmdEqnErrs[$24  = 2^3 
$ ]& \textcolor{red}{\xmark}& \textcolor{red}{\xmark}& \\\cdashline{2-6}[0.2pt/3pt]
&\cmdEqnErrs[$26  = 2^2 $]
 & \textcolor{red}{\xmark}& \textcolor{red}{\xmark}& \\\cdashline{2-6}[0.2pt/3pt]
&\cmdEqnErrs[$ \leadsto {\sigma_2} ({24}) =  {\sigma_2} ({26}) $]
  & \textcolor{red}{\xmark}& \textcolor{red}{\xmark}\\\cdashline{2-6}[0.2pt/3pt]
&$\blacksquare$ & \\
\end{tabular}
\scriptsize
\setlength{\tabcolsep}{2pt}
\centering
\begin{tabular}[t]{lp{8.5cm}cccccccccc}
\toprule
\textbf{Theorem 2} & \textbf{Trace of Sum of Matrices is Sum of Traces} \\
& Let $\mathbf A = [a]_n$ and $\mathbf B =  [b]_n$ be \underline{square matrices} of \underline{order} $n$. 
 
 Let $\mathbf A + \mathbf B$ denote the \underline{matrix entrywise sum} of $\mathbf A$ and $\mathbf B$.\\
&Then: 
 $ tr({\mathbf A + \mathbf B}) =  tr({\mathbf A}) +  tr({\mathbf B})$ 
 where $tr({\mathbf A})$ denotes the \underline{trace} of $\mathbf A$. \\
\midrule
& \textbf{Proof:} & correct& \cmdEqnErrs[eqn.]& \cmdRefErrs[ref.] \\
\mnamepp &  
 \cmdEqnErrs[$ tr( {\mathbf A + \mathbf B} )
       = \sum_{\mathbf C = \mathbf A + \mathbf B}  tr ({\mathbf C} )
        $]\quad\ \cmdRefErrs[$\underline{\text{Trace of Matrix}}$]
 & \textcolor{red}{\xmark}& \textcolor{red}{\xmark}&\textcolor{red}{\xmark}\\\cdashline{2-6}[0.2pt/3pt]
&\cmdEqnErrs[$ = \sum_{\mathbf C = \mathbf A}  tr( {\mathbf C}) + \sum_{\mathbf C = \mathbf B}  tr( {\mathbf C} )$]$\ $\cmdRefErrs[$\underline{\text{Sum of Summations equals Summation of Sum}}$]
        & \textcolor{red}{\xmark}& \textcolor{red}{\xmark}&  \textcolor{red}{\xmark}\\\cdashline{2-6}[0.2pt/3pt]
&\cmdEqnErrs[$ =  tr(\mathbf A) +  tr(\mathbf B)$]$
        \quad $\cmdRefErrs[$\underline{\text{Trace of Matrix}} $]
  & \textcolor{red}{\xmark}&\textcolor{red}{\xmark}& \textcolor{red}{\xmark}\\\cdashline{2-6}[0.2pt/3pt]
&$\blacksquare$ &\\
\bottomrule
\end{tabular}
\scriptsize
\setlength{\tabcolsep}{2pt}
\centering
\begin{tabular}[t]{lp{8.5cm}cccccccccc}
\toprule
\textbf{Theorem 3} & \textbf{Moment Generating Function of Discrete Uniform Distribution} \\

& Let $X$ be a \underline{discrete random variable} with a \underline{discrete uniform distribution with parameter $n$} 

\quad\quad for some $n \in \mathbb{N}$.\\

&Then the \underline{moment generating function} $M_X$ of $X$ is given by:  
 
\quad\quad $ {M_X}( t) = \dfrac {e^t (1 - e^{n t} ) } {n (1 - e^t) }$ \\
\midrule
& \textbf{Proof:} & correct& \cmdEqnErrs[eqn.]& \cmdRefErrs[ref.] \\
\mnamepp & From the definition of the \cmdRefErrs[$\underline{\text{discrete uniform distribution}}$], 

\quad$X$ has \underline{probability mass function}: \cmdEqnErrs[$\Pr ({X = k}) = \dfrac n {n + k}$] & \textcolor{red}{\xmark}& \textcolor{red}{\xmark}& \textcolor{red}{\xmark}\\\cdashline{2-6}[0.2pt/3pt]

&From the definition of a \underline{moment generating function}: 

\quad\quad${M_X} (t) = \mathbb{E} {e^{t X} } = \sum_{k  = 0}^n  \Pr ({X = k}) e^{t k}$ & \cmark& & \\\cdashline{2-6}[0.2pt/3pt]

&So: 

\quad\quad${M_X} (t) = \sum_{k \mathop = 0}^n  \Pr ({X = k}) e^{t k} $ & \cmark& & \\\cdashline{2-6}[0.2pt/3pt]
&\quad\quad\cmdEqnErrs[$= \frac n {n + k} \sum_{k \mathop = 0}^n ( {{(e^{t k} )}^n \prod_{j \mathop = 0}^{n - 1} ({e^{t \frac {j + 1} n} }) })$]\quad \cmdRefErrs[$\underline{\text{Sum of Geometric Sequence}}$] & \textcolor{red}{\xmark}& \textcolor{red}{\xmark}& \textcolor{red}{\xmark}\\\cdashline{2-6}[0.2pt/3pt]
&\quad\quad\cmdEqnErrs[$ = \frac n {n + k} \sum_{j \mathop = 0}^n ({e^{t \frac {j + 1} n} e^{t j} } )
$] & \textcolor{red}{\xmark}&  \textcolor{red}{\xmark}& \\\cdashline{2-6}[0.2pt/3pt]
&\quad\quad\cmdEqnErrs[$= \frac n {n + k}  ({1 - e^{n t} } )$]
 \quad \cmdRefErrs[$\underline{\text{Discrete Uniform Distribution}}$] & \textcolor{red}{\xmark}&  \textcolor{red}{\xmark}& \textcolor{red}{\xmark}\\\cdashline{2-6}[0.2pt/3pt]
&\quad\quad\cmdEqnErrs[$ = \frac {e^t ({1 - e^{n t} }) } {n ({1 - e^t}) } $]  & \textcolor{red}{\xmark}&  \textcolor{red}{\xmark}& \\\cdashline{2-6}[0.2pt/3pt]
& $\blacksquare$\\
\bottomrule
\end{tabular}
\captionof{table}{\textit{Equation-related errors} in full proof generation. \methodname can struggle with  \cmdEqnErrs[invalid equations] and \cmdEqnErrs[derivations], including  basic equalities (Theorem 1), and more sophisticated settings (Theorems 2 and 3).
Since derivations involve deploying definitions and rules from references and using references as justification, equation-related errors are often coupled with \cmdRefErrs[reference errors].
}
\label{tab:err_eqn}

}

\newpage
\subsection{Examples}
\label{sec:examples}

{\scriptsize
\setlength{\tabcolsep}{2pt}
\centering
\begin{tabular}[t]{p{2.2cm}p{4.5cm}p{2.1cm}p{4cm}}
\toprule
\textbf{Theorem} & \textbf{Metric Space is Perfectly Normal} & \textit{Gold Proof}\\

& Let $M =  ({A, d})$ be a \underline{metric space}. & \multicolumn{2}{l}{By definition, a \underline{topological space} is \underline{perfectly normal space} {{iff}} it is:  }\\
&Then $M$ is a \underline{perfectly normal space}. & \multicolumn{2}{l}{\quad \underline{perfectly $T_4$ space} and \underline{$T_1$ (Fréchet) space}.} \\
&&\multicolumn{2}{l}{We have that:}  \\
&&\multicolumn{2}{l}{\quad \underline{\cmdref[Metric Space is Perfectly $T_4$]}}\\
&&\multicolumn{2}{l}{\quad \underline{\cmdreff[Metric Space is $T_2$ (Hausdorff)]}}\\ 
&&\multicolumn{2}{l}{\quad \textcolor{blue}{\underline{$T_2$ (Hausdorff) Space is a $T_1$ (Frechet) Space}}.}\\

&&$\blacksquare$ \\
\midrule
$\methodnamenospace_{++}$ & From:  

\quad \underline{\cmdreff[Metric Space is Hausdorff]} & $\methodnamenospace$ \textit{(without stepwise++)} & From:
 
 \quad \underline{\cmdref[Metric Space is Perfectly $T_4$]} \\
 
 &\quad \textcolor{blue}{\underline{$T_2$ (Hausdorff) Space is $T_1$ Space}}  
 
 \quad \underline{\cmdref[Metric Space is Perfectly $T_4$]} 
 
 it follows that $M$ is a \underline{topological space} which is \underline{perfectly normal}. & &\quad \underline{\cmdreff[Metric Space is $T_2$ (Hausdorff)]} 
 
\cmdOtherErrs[it follows that $M$]

\quad \cmdOtherErrs[ is a]\underline{\cmdOtherErrs[perfectly normal space]}.\\
\bottomrule
\end{tabular}
\vspace{2pt}
\captionof{table}{Example proof using provided in-context reference constraints. 
    The key theorem \texttt{$T_2$ Space is $T_1$ Space} is provided as a constraint, but under greedy decoding the model ignores the constraint, resulting in \cmdOtherErrs[skipping steps].
    Stepwise++ decoding selects proof steps based on likelihood and constraint satisfaction, resulting in better reference coverage and a correct proof.
}
\label{tbl:example}
}

{
\begin{tabular}{p{3cm} l}
\toprule
\textbf{Theorem Title} & \texttt{Equality of Complex Numbers} \\
\textbf{Theorem Content} & \includegraphics[scale=0.5]{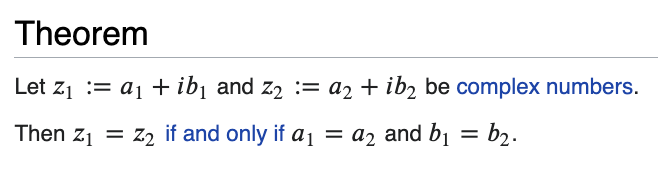} \\
\midrule
Gold Proof & \includegraphics[scale=0.5]{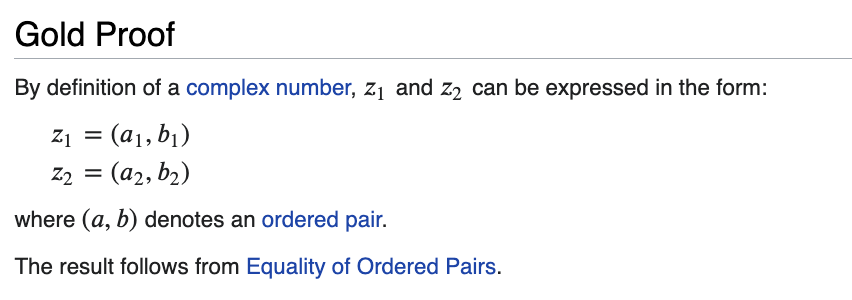} \\
\midrule
\methodnamenospace & \includegraphics[scale=0.4]{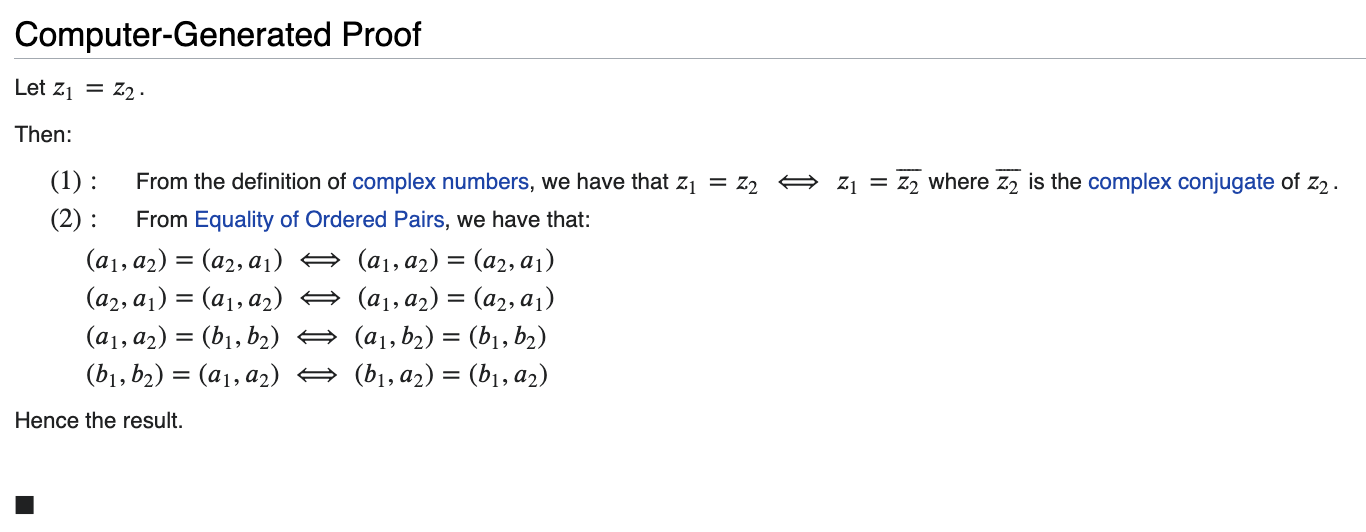} \\
\midrule
\mnamepp & \includegraphics[scale=0.5]{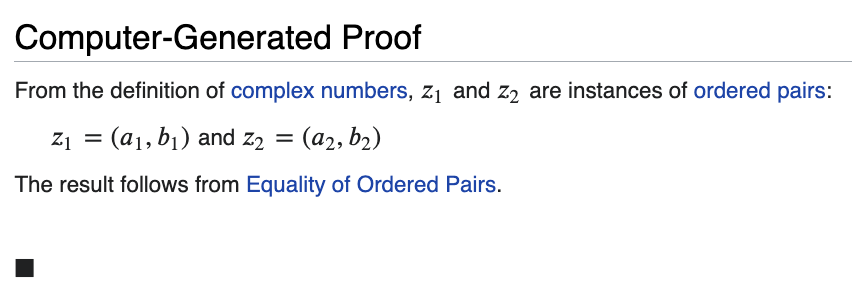} \\
\bottomrule
\end{tabular}
\vspace{2pt}
\captionof{table}{
    A \textit{complex}, direct proof.
    Without stepwise++ decoding, \methodname makes an \textit{invalid deployment} error, continues with some nonsense, and prematurely terminates the proof.
    The \mnamepp proof is correct, thanks to stepwise++ decoding.
}
\label{tab:qual_complex_direct}
}
{
\begin{tabular}{p{3cm} l}
\toprule
\textbf{Theorem Title} & \texttt{Compact Complement Topology is Connected} \\
\textbf{Theorem Content} & \includegraphics[scale=0.5]{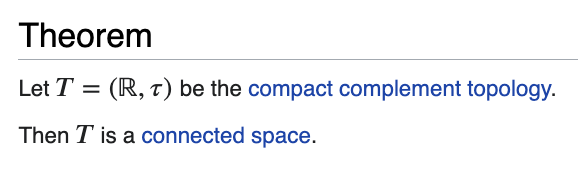} \\
\midrule
Gold Proof & \includegraphics[scale=0.5]{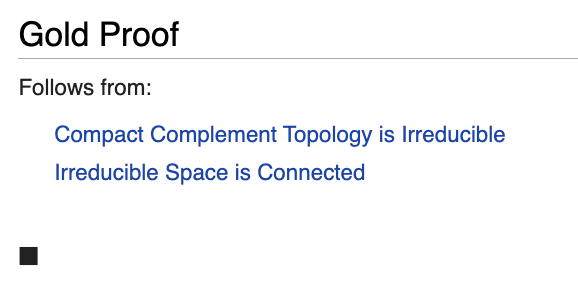} \\
\midrule
GPT-3 & \includegraphics[scale=0.5]{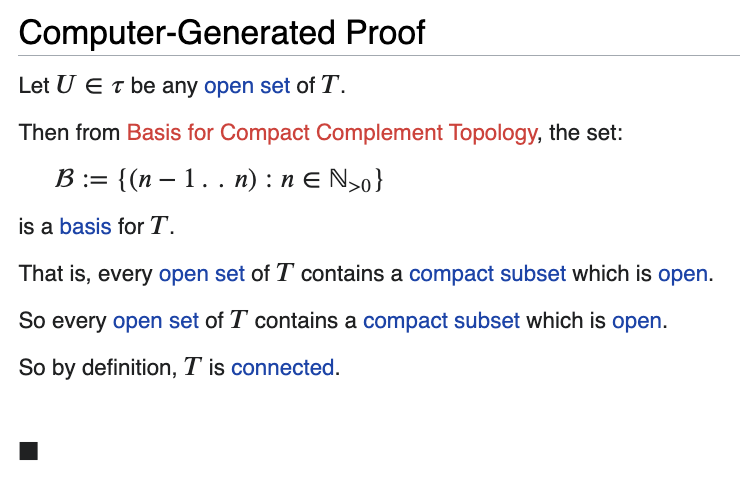} \\
\midrule
\methodname \textsc{Retrieve} & \includegraphics[scale=0.5]{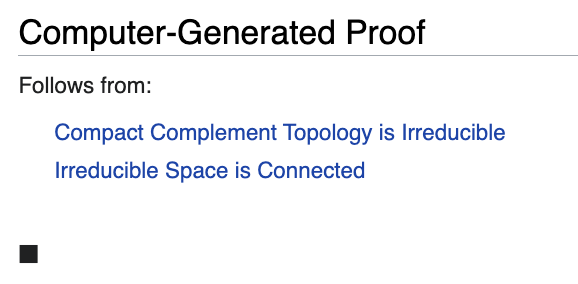} \includegraphics[scale=0.3]{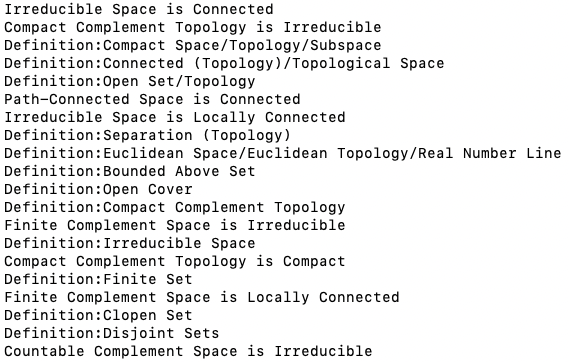} \\
\bottomrule
\end{tabular}
\vspace{2pt}
\captionof{table}{
    A \textit{reference assembly} proof.
    GPT-3's proof is incorrect, possibly because it doesn't know to use the two references.
    $\methodname_{\textsc{Retrieve}}$ uses retrieved references (shown on the right) to arrive at a correct proof.
}
\label{tab:qual_reference}
}
{
\begin{tabular}{p{2.6cm} l}
\toprule
\textbf{Theorem Title} & \texttt{Pointwise Addition on Real-Valued Functions is Associative} \\
\textbf{Theorem Content} & \includegraphics[scale=0.5]{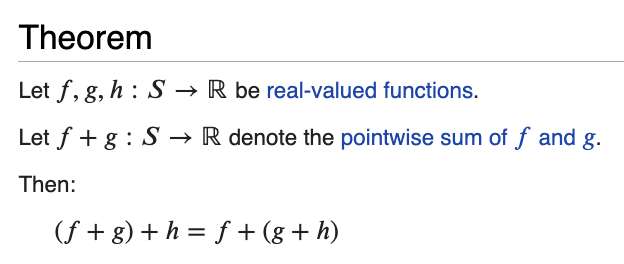} \\
\midrule
Gold Proof & \includegraphics[scale=0.5]{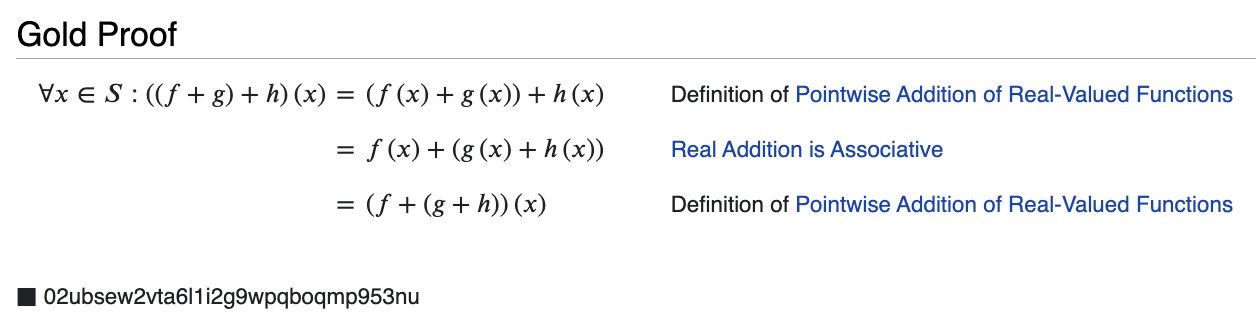} \\
\midrule
GPT-3 & \includegraphics[scale=0.5]{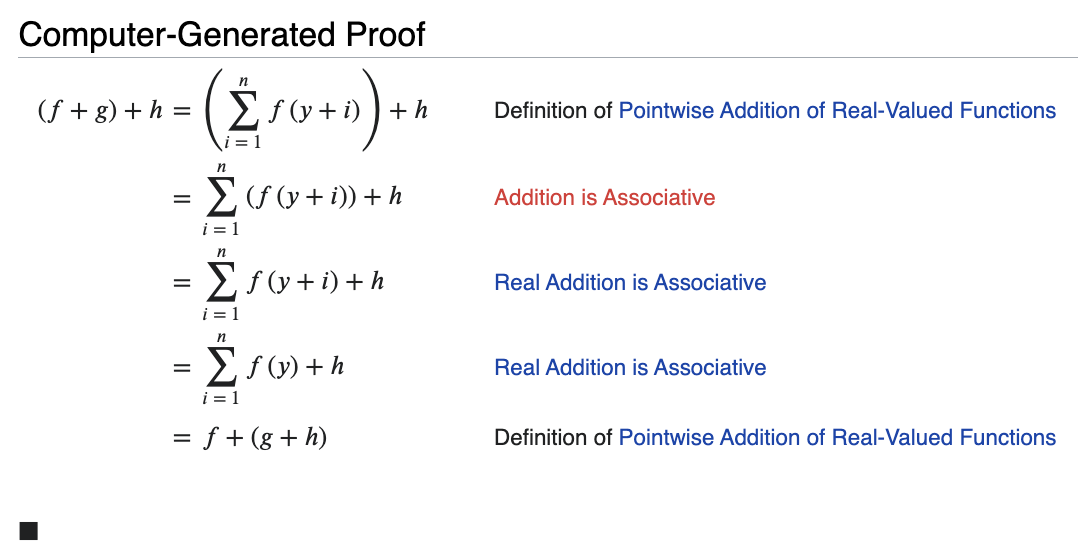} \\
\midrule
\methodname & \includegraphics[scale=0.5]{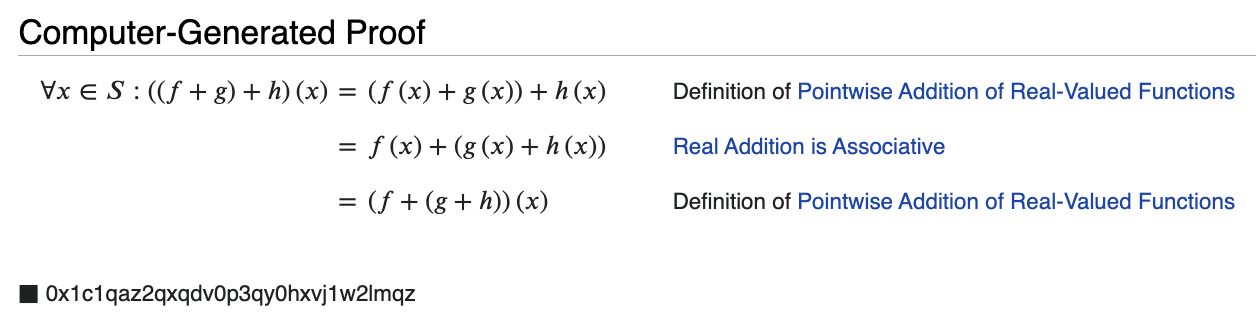} \\
\bottomrule
\end{tabular}
\vspace{2pt}
\captionof{table}{
    A \textit{template adaptation} proof, which is proved via symbolic derivations.
    \methodname adapts the proof of a similar training theorem, \texttt{Pointwise Addition on Complex-Valued Functions is Associative}, to prove the claim.
    Despite training on the same (theorem, proof) pairs, vanilla GPT-3 fails to prove the claim.
}
\label{tab:qual_template_equations}
}
{\begin{tabular}{p{2.6cm} l}
\toprule
\textbf{Theorem Title} & \texttt{Cosine in terms of Sine} \\
\textbf{Theorem Content} & \includegraphics[scale=0.5]{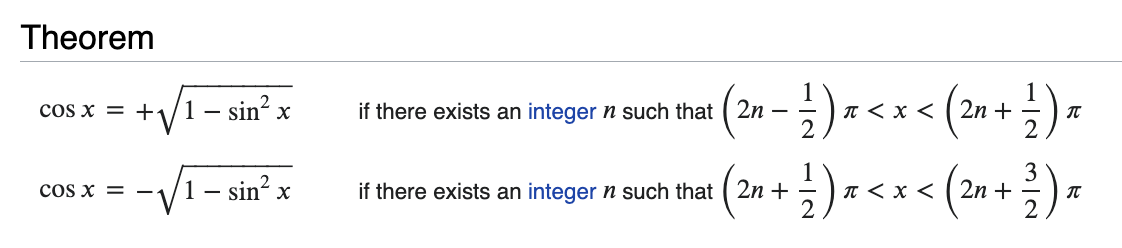} \\
\midrule
Gold Proof & \includegraphics[scale=0.5]{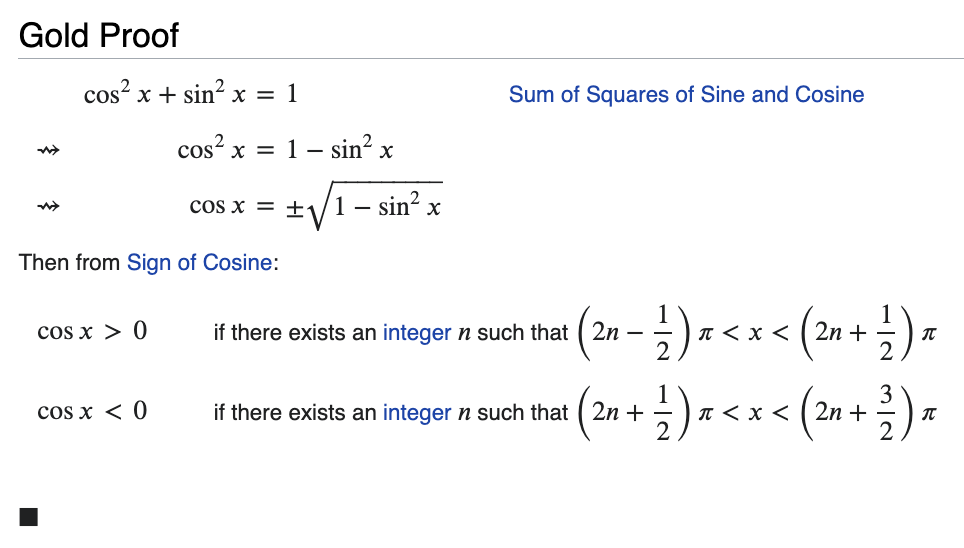} \\
\midrule
GPT-3 & \includegraphics[scale=0.35]{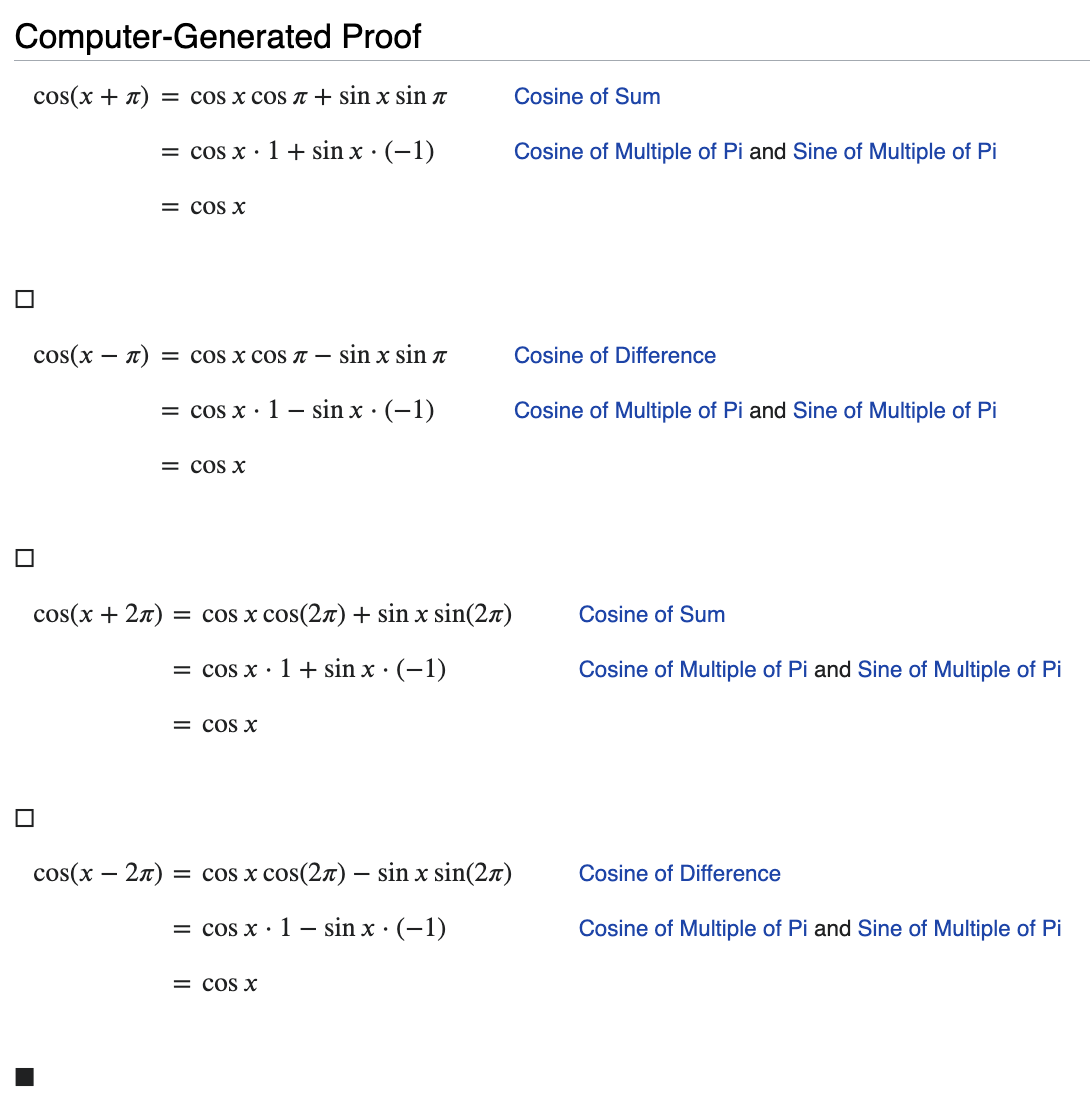} \\
\midrule
\methodname & \includegraphics[scale=0.5]{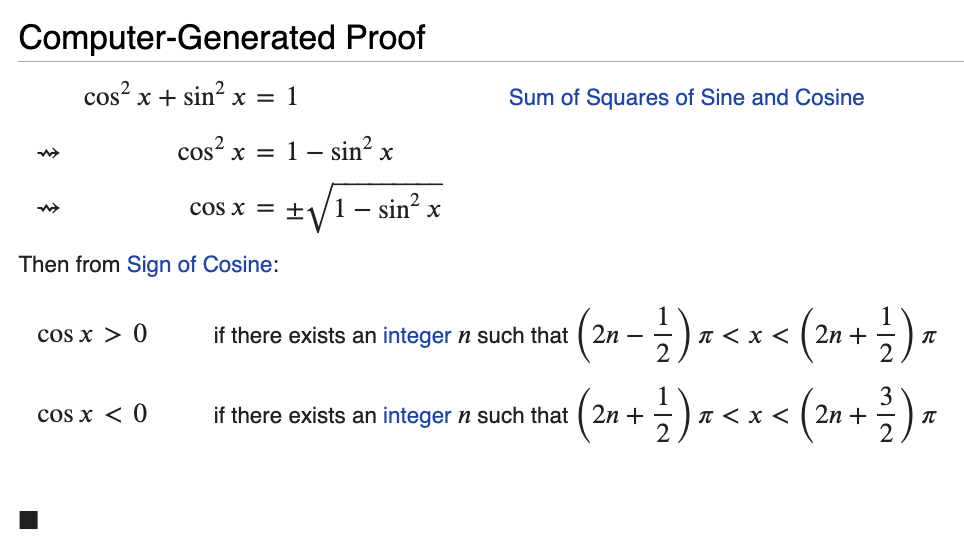} \\
\bottomrule
\end{tabular}
\captionof{table}{
    A \textit{template adaptation} proof by cases.
    GPT-3's proof goes completely derailed and it does not know to use the reference \texttt{Sum of Squares of Sine and Cosine}.
    \methodnamenospace's proof is correct.
    The model adapts the proof of the mirroring theorem, \texttt{Sine in terms of Cosine}, in the training set.
}
\label{tab:qual_template_cases}
}
{
\begin{tabular}{c c}
\toprule
\multicolumn{2}{l}{\textbf{Theorem Title:} \texttt{Triangle Inequality/Complex Numbers/General Result}} \\
\multicolumn{2}{l}{\textbf{Theorem Content:} \includegraphics[scale=0.5]{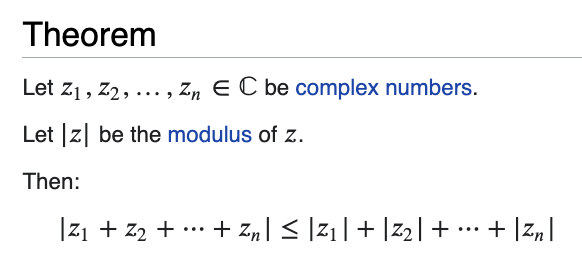}} \\
\midrule
Gold Proof & \methodnamenospace \\
\includegraphics[scale=0.33]{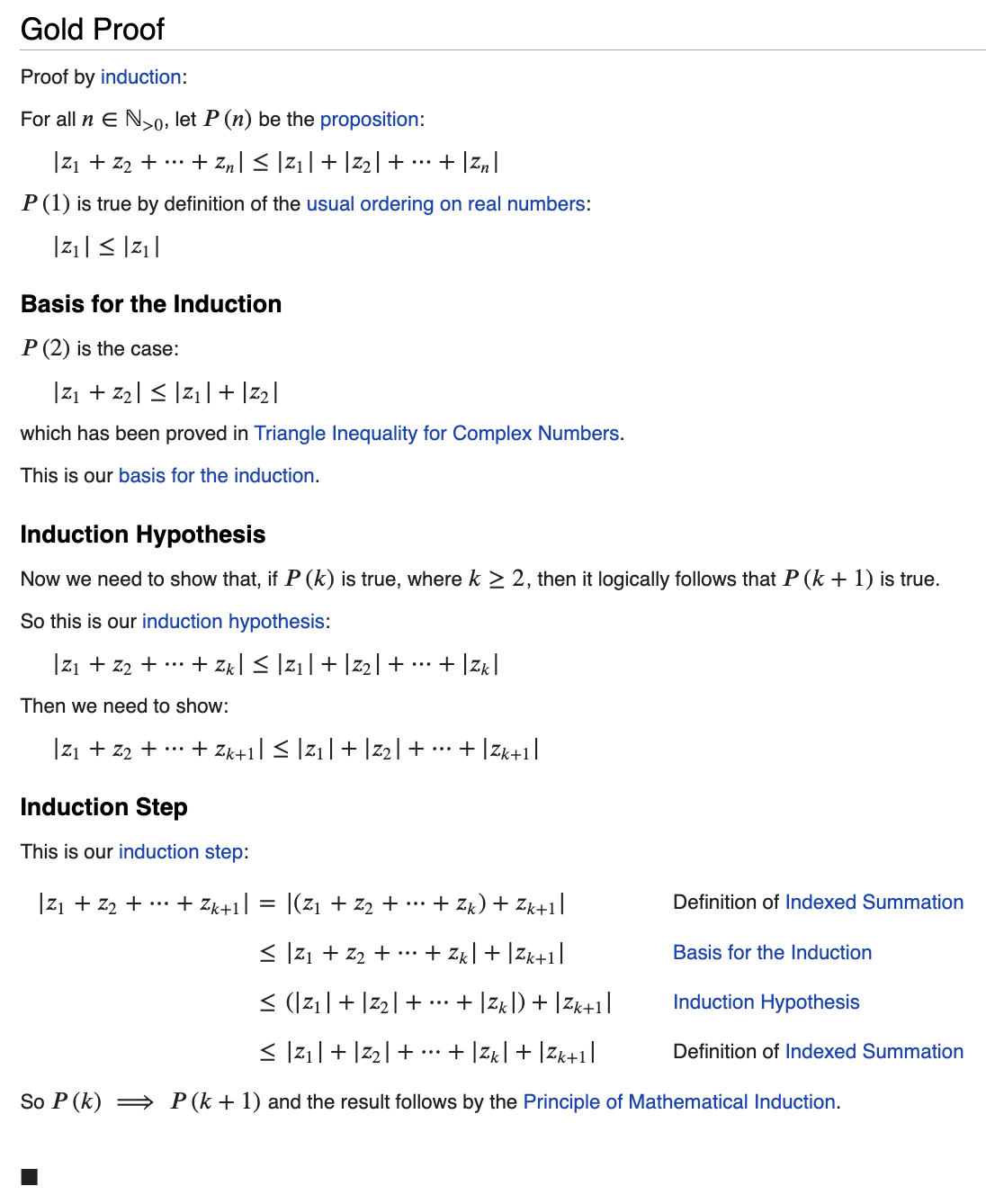} & \includegraphics[scale=0.33]{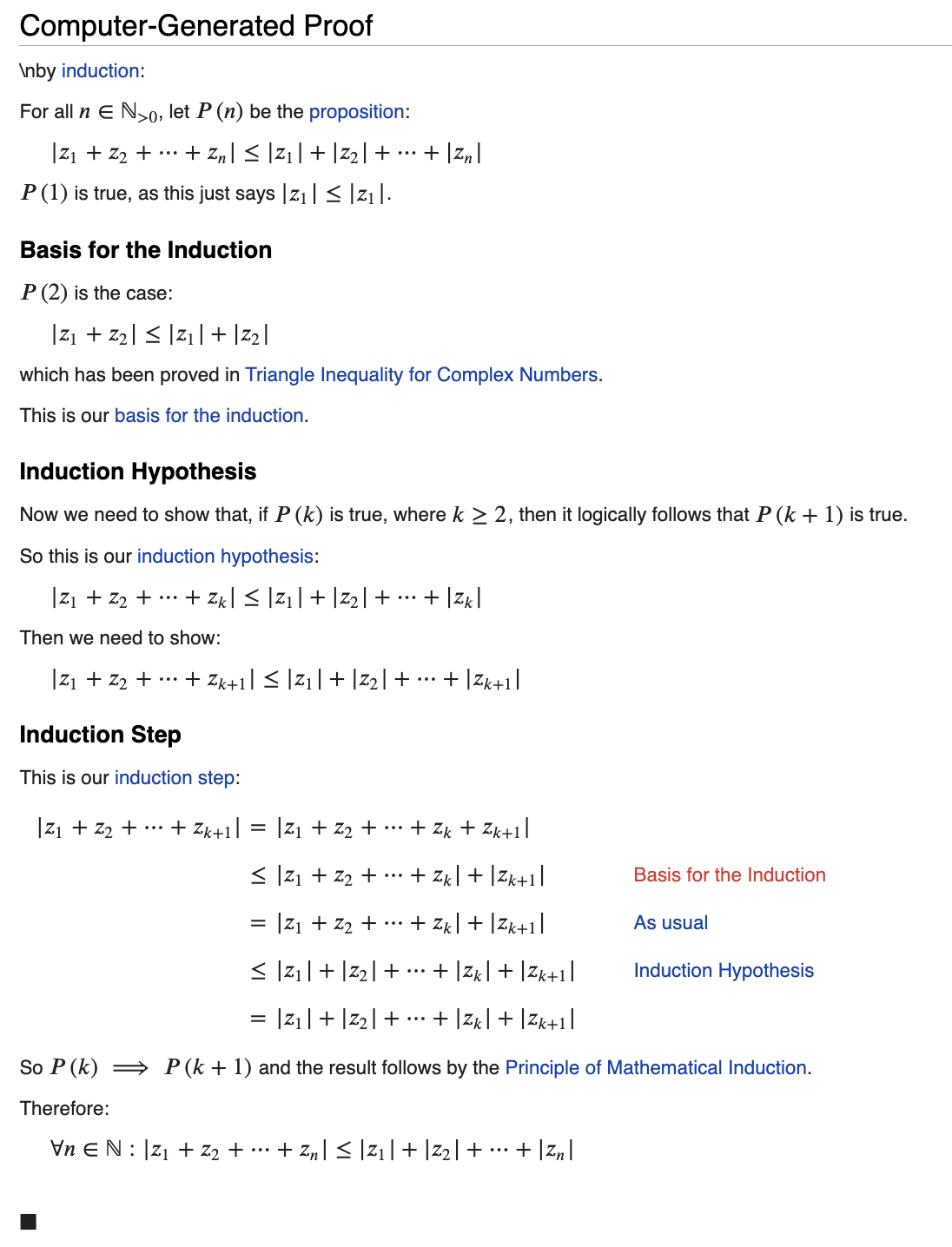} \\
\bottomrule
\end{tabular}
\captionof{table}{
    A \textit{complex} proof by induction.
    \methodnamenospace's proof makes one \cmdHalluc[hallucinated reference] error, one \cmdRepetition[repetition] error, and is otherwise correct.
    The model did not see a similar proof during training: while there are more variants of the \texttt{Triangle Inequality} theorem in our dataset (i.e. with \texttt{Real Numbers} and \texttt{Geometry}), they only discuss the 2-variable case and none of them discuss the $n$-variable general result.
    So in this case, the model has learned the format of proof by induction and can apply it in new context.
    (A proof-by-induction example in train set: \texttt{Sum of Sequence of Squares/Proof by Induction}.)
}
\label{tab:qual_complex_induction}
}

\newpage

\section{Dataset Details}
\label{sec:dataset}

We provide an overview of \textsc{NaturalProofs} and its ProofWiki domain from which we build \textsc{NaturalProofs-Gen}.
Refer to \citep{welleck2021naturalproofs} for further details about \textsc{NaturalProofs}.

Our dataset is derived from $\natproofs$, a multi-domain corpus of theorem statements, proofs, definitions, and additional pages (e.g. axioms, corollaries) in natural mathematical language.
We use the ProofWiki\footnote{The ProofWiki domain of \natproofs{} dataset is under the CC BY-SA 4.0 license.} domain, which provides broad-coverage of many subject areas (e.g. Set Theory, Analysis) sourced from ProofWiki, an online compendium of community-contributed mathematical proofs.
$\textsc{ProofWiki}$ contains $\sim$20k theorems, $\sim$20k proofs, $\sim$12k definitions, and $\sim$1k additional pages (e.g. axioms, corollaries).
The set of all $\sim$33k theorems, definitions, and additional pages form the \textit{reference set} $\mathcal{R}$.
Finally, $\sim$14.5k of the theorems $\xb$ are paired with at least one proof $\yb$ to form \textit{examples} $\mathcal{D}=\{(\xb,\yb)_i\}_{i=1}^N$.
\cite{welleck2021naturalproofs} split the reference sets and examples into training, validation, and test splits to ensure that no theorem in the validation or test splits was mentioned in the training split.

\section{Segment-level Constrained Decoding}
\label{sec:expr_decoding}

In this section we present a generic segment-level decoding algorithm that contains stepwise++, full-proof sampling, and greedy decoding as special cases. 
We generate a multi-step proof using a value function $v(\cdot)$ that measures language quality and constraint satisfaction.
Search can be done at the step-level, in which candidate next-steps are generated and high-value steps are retained in a beam, or at the proof-level, in which multiple proofs are generated and the highest-value proof is selected.
We formalize these into a generic \textit{segment-level search}, where a segment $s_t$ is either a proof-step $y_t$ or a full proof $\yb$.

The search iteratively builds a multi-step proof $\yb=(y_1,\ldots,y_T)$ by \textit{expanding}, \textit{scoring}, and \textit{selecting} a set of candidate \textit{segments}:
\begin{itemize}
    \item \texttt{Expand}$: S_{t-1}\rightarrow S_t'$ extends segments $S_{t-1}=\{s_{\leq t}\}$ into candidates $S_{t}'=\{(s_{\leq t}, s_t)\}$.
    \item \texttt{Score} : $(s_{\leq t}, v)\rightarrow \mathbb{R}$  scores a candidate using a value function, $v(s_{\leq t})\rightarrow \mathbb{R}$.
    \item \texttt{Select} : $S_t'\rightarrow S_t$ prunes  candidates $S_t'$ into segments $S_t$ used in the next iteration.
\end{itemize}

\paragraph{Value function.} We score candidates based on constraint satisfaction and language quality,
\begin{align}
    v(s_{\leq t}) = \alpha v_{\text{constraint}}(s_{\leq t}) + (1-\alpha)v_{\text{LM}}(s_{\leq t}),
\end{align}
where $v_{\text{constraint}}(y_{\leq t})$ is the number of unique in-context reference-titles in $s_{\leq t}$, and $v_{\text{LM}}(s_{\leq t})$ is $\log p_\theta(s_{\leq t})$. We normalize each term by dividing by the maximum absolute value among candidates.

\paragraph{Greedy search.} This baseline search defines a \texttt{segment} as a full proof, meaning $s_0$ is an empty sequence and $s_1$ is a proof $\yb$. \texttt{Expand} samples one segment candidate with temperature 0. \texttt{Score} and \texttt{select} are trivial since there is only one candidate.
Greedy search costs $T$ steps of tokens.

\paragraph{Sample-and-rerank.} In this search, a \texttt{segment} is again full proof, but \texttt{expand} samples \textit{N} candidates, $S_1'=\{\yb^{n}\sim q(\cdot|\xb)\}_{n=1}^N$, where $q$ is a decoding algorithm (e.g. temperature sampling). \texttt{Select} takes the top scoring candidate, $\yb = \argmax_{\yb^{n}\in S_1'} v(\yb^{n})$.
The cost is $NT$ steps of tokens.

\paragraph{Step-wise stochastic beam search.} This search generates by iteratively sampling and re-ranking next-step candidates.
In this case, a segment is a proof step, $y_t$, and 
each iteration starts with a beam of proofs-so-far, $S_{t-1}=\{y_{< t}^{k}\}_{k=1}^K$, where $K$ is the beam size.
 \texttt{Expand} samples $N$ next-step candidates for each proof-so-far in the beam,
\begin{align}
\label{eqn:expand-apx}
S_t'=\bigcup_{y_{<t}\in S_{t-1}}\big\{ (y_{<t} \circ y^{n}_{t})\ |\ y^n_t\sim q(\cdot |y_{< t},\xb)\big\}_{n=1}^N,
\end{align}
where $q$ is a decoding algorithm (e.g. temperature sampling) and $\circ$ is concatenation.  \texttt{Select} forms the next beam using the top-$K$ scoring candidates,
\begin{align}
\label{eqn:select}
    S_{t}=\argtopK_{y_{\leq t}\in S_t'}\ v(y_{\leq t}).
\end{align}
When a proof in the beam terminates, it is not expanded further. The search ends when the beam consists of $K$ terminated proofs.
The highest scoring proof is returned as the final output.
The cost is $NTK$ steps of tokens.

\paragraph{Stepwise++.} 
At certain proof steps it is important to enumerate and explore options, while at others (e.g. derivations) a single highly probable prediction is better.
To this end, we \texttt{expand} by sampling with multiple temperatures, meaning that we expand each prefix $y_{<t}$ in (\ref{eqn:expand}) using:
\begin{align}
    \{y_t^n\sim q_\tau(\cdot|y_{<t},\xb)\ |\ \tau\in \{\tau_1,\ldots, \tau_m\}\},
\end{align}
where $q_\tau$ is sampling with temperature $\tau$.
This relaxes the commitment to a single temperature for all proof steps, intuitively balancing exploration (higher $\tau$) with exploitation (lower $\tau$).

Second, during the search we want to balance selecting proof steps that satisfy constraints and proof steps with high log-probability.
To this end, we \texttt{select} clusters with different value weights,
\begin{align}
    S_t=\{y_{\leq t} \in \mathrm{top}_{K'} (S_\alpha)\ |\ \alpha\in \{\alpha_1,\ldots,\alpha_\ell\}\},
\end{align}
where $S_\alpha$ means the set of candidates scored with $v=\alpha v_{\text{constraint}}+(1-\alpha)v_{\text{LM}}$, and $K'=K/\ell$.
This interpolates between selecting steps with good language score ($\alpha$ small), constraint score ($\alpha$ large), and balance ($\alpha:$ 0.5).

\section{Implementation Details and Experimental Setup}
\label{sec:implementation}

\paragraph{Data preprocessing.} We automatically infer the boundaries of proof steps within the raw proof contents, and merge contiguous lines into atomic proof steps when appropriate.
Steps are separated by the \texttt{\textbackslash{}n} token (\texttt{\textbackslash{}\textbackslash{}n} in Python string), and lines within a step are separated by the newline token (\texttt{\textbackslash{}n} in Python string).

\paragraph{Additional model details.}
All GPT-3 models (including \methodnamenospace{} models) are fine-tuned instances of the \texttt{Curie} engine, the second largest model available through the OpenAI API at the time of writing.\footnote{https://beta.openai.com/docs/guides/fine-tuning}
The model's performance on the EleutherAI evaluation harness\footnote{\url{https://github.com/EleutherAI/lm-evaluation-harness}} is between the 6.7B and 13B variants of the autoregressive transformer language model GPT-3 from \citep{brown2020gpt3},\footnote{\url{https://blog.eleuther.ai/gpt3-model-sizes/}} though further details of the Curie model are not publicly available.

Separately, we fine-tune GPT-J 6B,\footnote{\url{https://huggingface.co/EleutherAI/gpt-j-6B}} a publicly available autoregressive transformer language model trained on the Pile~\citep{pile}, GPT-2~\citep{radford2019language}, an autoregressive transformer language model trained on scraped web documents, and GPT-Neo-125M,\footnote{\url{https://github.com/EleutherAI/gpt-neo}} a GPT-2 like causal language model trained on the Pile.

Our retrieval model is the joint retrieval model from \citep{welleck2021naturalproofs} trained for reference retrieval on ProofWiki using the same dataset splits as NaturalProver.
We use the publicly-available pretrained model from the GitHub repository of \citep{welleck2021naturalproofs} and do not update the model further.
We use the model to retrieve the top-20 references for each input theorem.

\paragraph{Implementation details.}
All GPT-3 models (including \methodnamenospace{} models) are fine-tuned with the OpenAI API\footnote{\url{https://beta.openai.com/docs/guides/fine-tuning}} for 4 epochs with a batch size of 64.
Other models (GPT-2/J/Neo) are trained on one Quadro RTX 8000 GPU.
During inference, the prompt (up to \texttt{<proof>}) is truncated to 1024 tokens.
For full proof generation, we allow a maximum of 1020 generated tokens.
For next-step suggestion, we truncate the proof-so-far to 900 tokens, and allow a maximum of 120 generated tokens per step.

\paragraph{Stepwise++ decoding.}
For expansion with multiple temperatures, we use $N=10$ candidates sampled with $(n,\tau)\in \{(1,0.0), (3,0.3), (3,0.5), (3,0.7)\}$.
We also tried including $\tau=1.0$ which resulted in very poor $\textsc{Gleu}$, and \{(1,0.0), (5,0.3), (4,0.5)\}.
For selection, we use a beam size $K=9$, and three equally-sized clusters formed with $\alpha\in \{0.1,0.5,1.0\}$. We also tried $\{0.5,0.75,0.9\}$.
We use $\alpha=0.75$ to pick select the final sequence, based on our ablation with full-proof sampling.

\paragraph{Full proof sampling.}
We use temperature $\tau=0.3$, selected based on a search over $\tau\in\{0.1,0.3,0.5,0.7\}$ using GLEU plus Ref-F1 on the core dev set.

\section{Additional Evaluation Details}
\label{sec:additional_evaluation}

\subsection{Full Evaluation Schema}
\label{sec:full_schema}

\renewcommand{\arraystretch}{1} 
\begin{table}
\centering
\small
\begin{tabularx}{\textwidth}{lX}
\toprule
\textbf{Aspect / Error Type} & \textbf{Definition} \\
\midrule
\multicolumn{2}{c}{\textsc{Overall Evaluation}} \\
\midrule
\cmdOverallCorrect[\bf Correctness] & Choose a rating below. Not every statement in each rating will apply to the proof given the rating, but many statements will apply, and the general theme of the rating will hold: \\
& $\circ$ 0: The proof is missing. \\
& $\circ$ 1: The proof makes no sense or is unrelated to the problem statement. \\
& $\circ$ 2: The proof contains serious logical flaws and lacks adequate justification or explanation. \\
& $\circ$ 3: The proof has some gaps in reasoning. \\
& $\circ$ 4: The proof is correct or nearly correct and logically coherent. \\
& $\circ$ 5: The proof is correct and flows logically. \\
\midrule
\cmdOverallUseful[\bf Usefulness] & Even if the proof is not perfect, would it be useful to you if you were to prove this theorem? \\
& $\circ$ 0: The proof is missing. \\
& $\circ$ 1: Seeing this proof would not help with proving the theorem by myself at all. \\
& $\circ$ 2: Seeing this proof proof would slightly decrease the effort needed to prove the theorem by myself. \\
& $\circ$ 3: Seeing this proof would make it substantially easier to prove the theorem by myself. \\
& $\circ$ 4. The proof is almost correct, and only needs a few minor corrections. \\
& $\circ$ 5: The proof is correct and could be directly used as a solution. \\
\midrule
\multicolumn{2}{c}{\textsc{Step-wise Evaluation}} \\
\midrule
\cmdStepCorrect[\bf Correctness] & Is this step correct? \\
& $\circ$ Yes \\
& $\circ$ No (check this if you identified any error in previous questions) \\
& $\circ$ Cannot determine (e.g. this step makes a valid progress, but it depends on an invalid prior step) \\
& $\circ$ This is a meaningless step (e.g. QED) \\
\midrule
\cmdStepUseful[\bf Usefulness] & Could this step be a helpful hint for proving the theorem by myself? \\
& $\circ$ Yes \\
& $\circ$ No \\
\midrule
\cmdRefErrs[\bf Reasoning: Reference] \\ 
\hspace{1em} \cmdDeploy[Invalid Deployment] & A statement deployed from a reference is not consistent with the reference. \\
\hspace{1em} \cmdJustify[Invalid Justification] & A reference is used as invalid justification for a statement. \\
\hspace{1em} \cmdHalluc[Hallucinated Ref.] & A reference that does not exist is used. \\
\hspace{1em} \cmdLoop[Self Loop] & The step refers to the theorem itself. \\
\midrule
\cmdEqnErrs[\bf Reasoning: Equation] & \\
\hspace{1em} \cmdEquation[Invalid Equation] & A standalone equation or initial equation in a derivation is invalid. \\
\hspace{1em} \cmdDerivation[Invalid Derivation] & An equation in a derivation does not follow from the preceding steps. \\
\midrule
\cmdOtherErrs[\bf Reasoning: Other] & \\
\hspace{1em} \cmdSkip[Skips Steps] & The step assumes unproven statements, or skips non-trivial steps. \\
\hspace{1em} \cmdRepetition[Repetition] & The step is merely a repetition of known things. \\
\hspace{1em} \cmdInvalid[Invalid (Other)] & The step's reasoning is invalid for reasons not captured by the other categories. \\
\midrule
\cmdLangErrs[\bf Language] & \\
\hspace{1em} \cmdIncomplete[Incomplete] & The step is not a complete mathematical statement or equation. \\
\hspace{1em} \cmdMisformat[Misformatted Math] & A math expression is not properly formatted. \\
\hspace{1em} \cmdUnk[Unknown] & There is a mis-spelled word, or unrecognized math symbol. \\
\midrule
\cmdSymErrs[\bf Symbolic] & \\
\hspace{1em} \cmdUndefined[Undefined] & One of the symbols is undefined. \\
\hspace{1em} \cmdOverloaded[Overloaded] & One of the symbols has overloaded meanings. \\
\hspace{1em} \cmdMistyped[Mistyped] & A symbol usage is not well-typed. \\
\hspace{1em} \cmdUnconventional[Unconventional] & Unconventional notation is used. \\
\bottomrule
\end{tabularx}%
\vspace{2pt}
\caption{
    Detailed description of the human evaluation schema.
}
\label{tab:schema_definitions}
\end{table}
\renewcommand{\arraystretch}{1} 

\autoref{tab:schema_definitions} shows the full schema of human evaluation.
The overall \cmdOverallCorrect[correctness] and \cmdOverallUseful[usefulness] are rated on a 0-5 scale.
The step-wise \cmdStepCorrect[correctness] and \cmdStepUseful[usefulness] are yes/no questions, while the error types ask for a binary indicator for the existence of each error type.

\subsection{Additional Human Evaluation Details}
\label{sec:additional_human}

\paragraph{Process.}
The authors conducted and moderated group sessions with the annotators.
Each session consisted of 30-minutes of training and a 1-hour working/Q\&A period.
After attending the session, annotators could continue working on their assigned tasks for two weeks.
Each annotator was assigned 25 theorems (with 5 proofs per theorem, equaling 125 total tasks) and asked to complete as many tasks as they would like.
The evaluation guideline that the annotators referenced to can be found in the supplementary materials.
The pre-recorded training video is available at 
\url{https://drive.google.com/file/d/1TRS5XRf_coLEkC4lqaizaqSwHHgBPrG2}.

\begin{figure}
\centering
\begin{minipage}{0.83 \textwidth}
    \includegraphics[width=\textwidth]{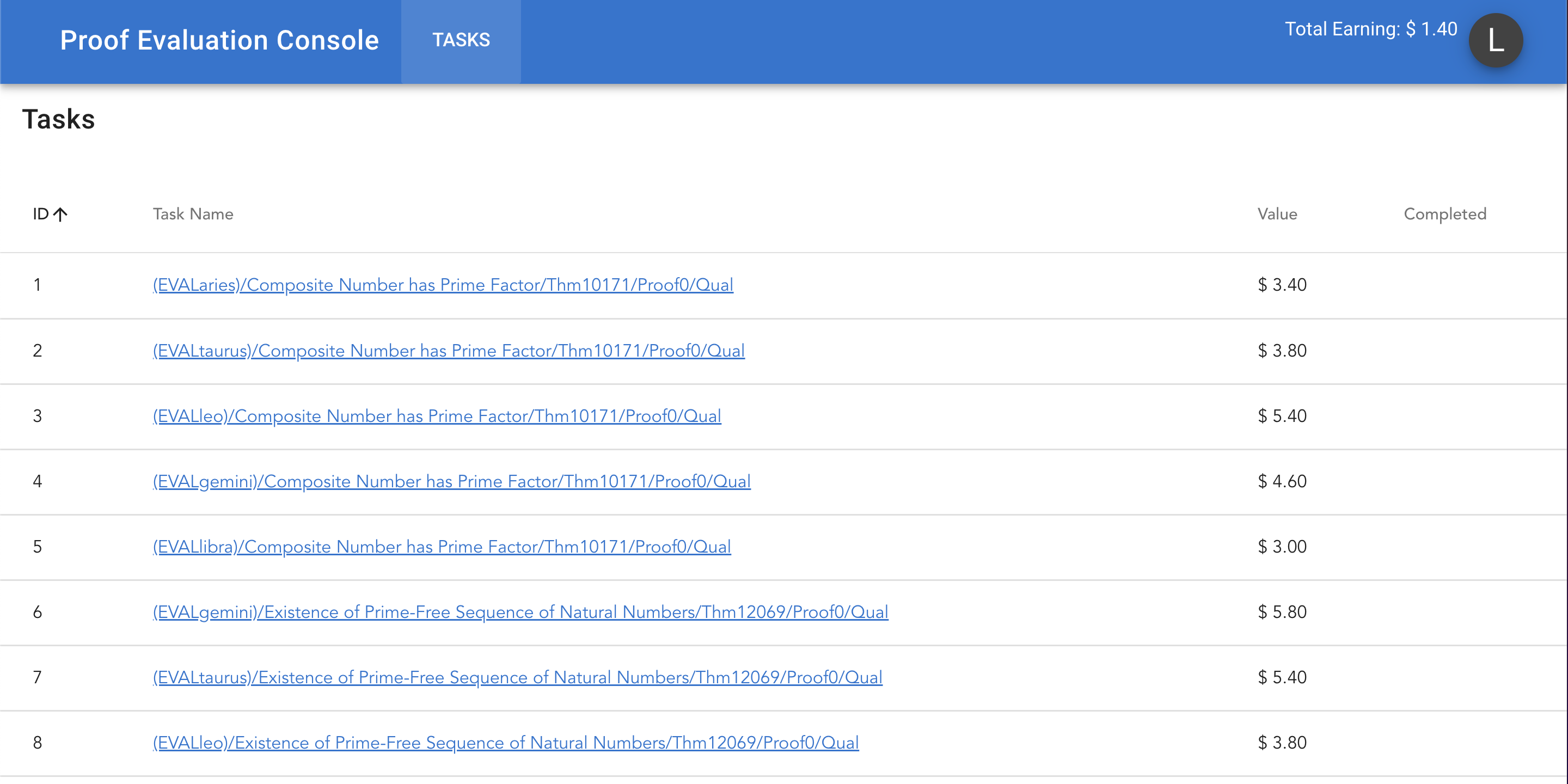}
\end{minipage}
\vspace{16pt}
\begin{minipage}{0.4 \textwidth}
    \includegraphics[width=\textwidth]{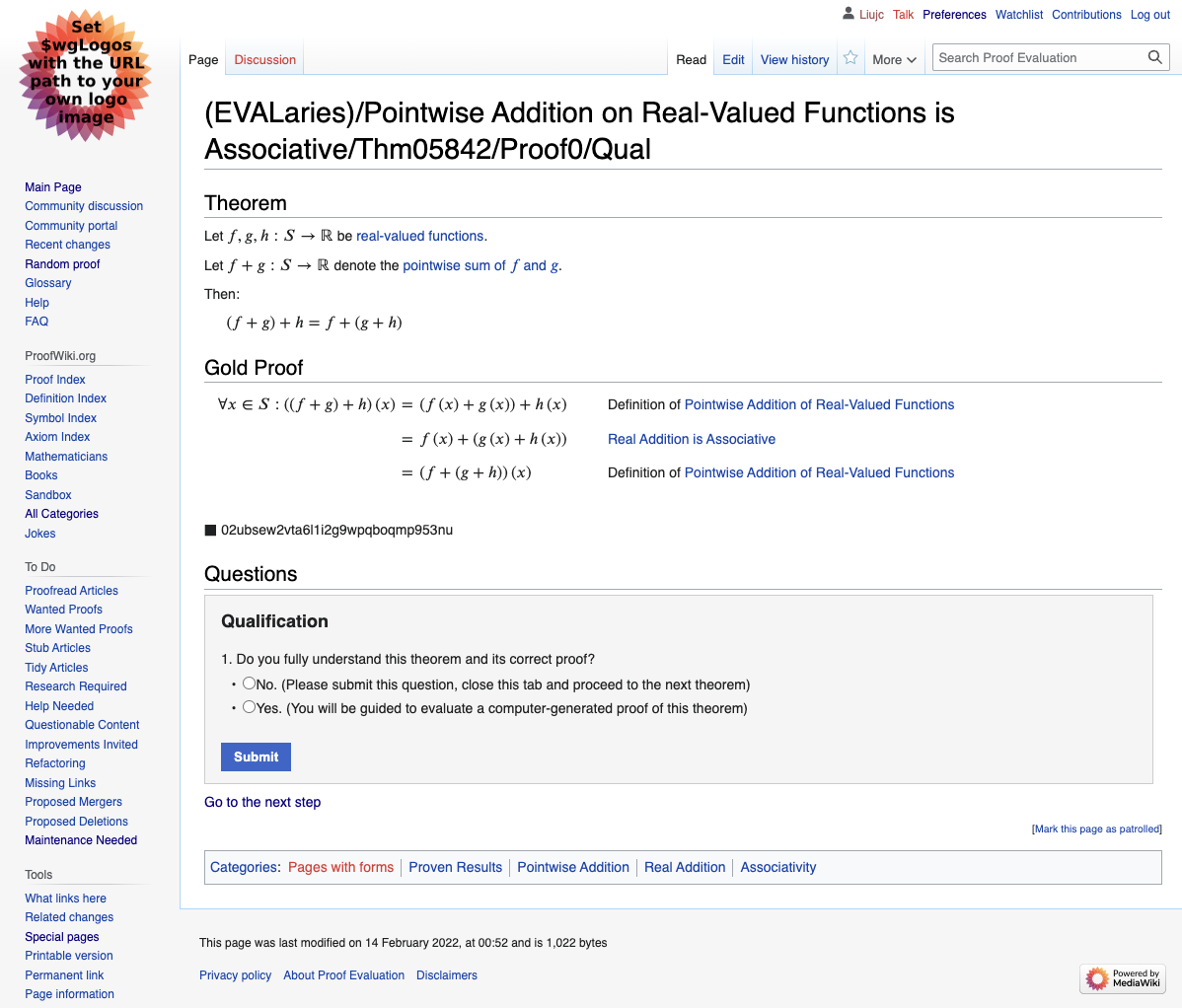}
    \vspace{16pt}
    \includegraphics[width=\textwidth]{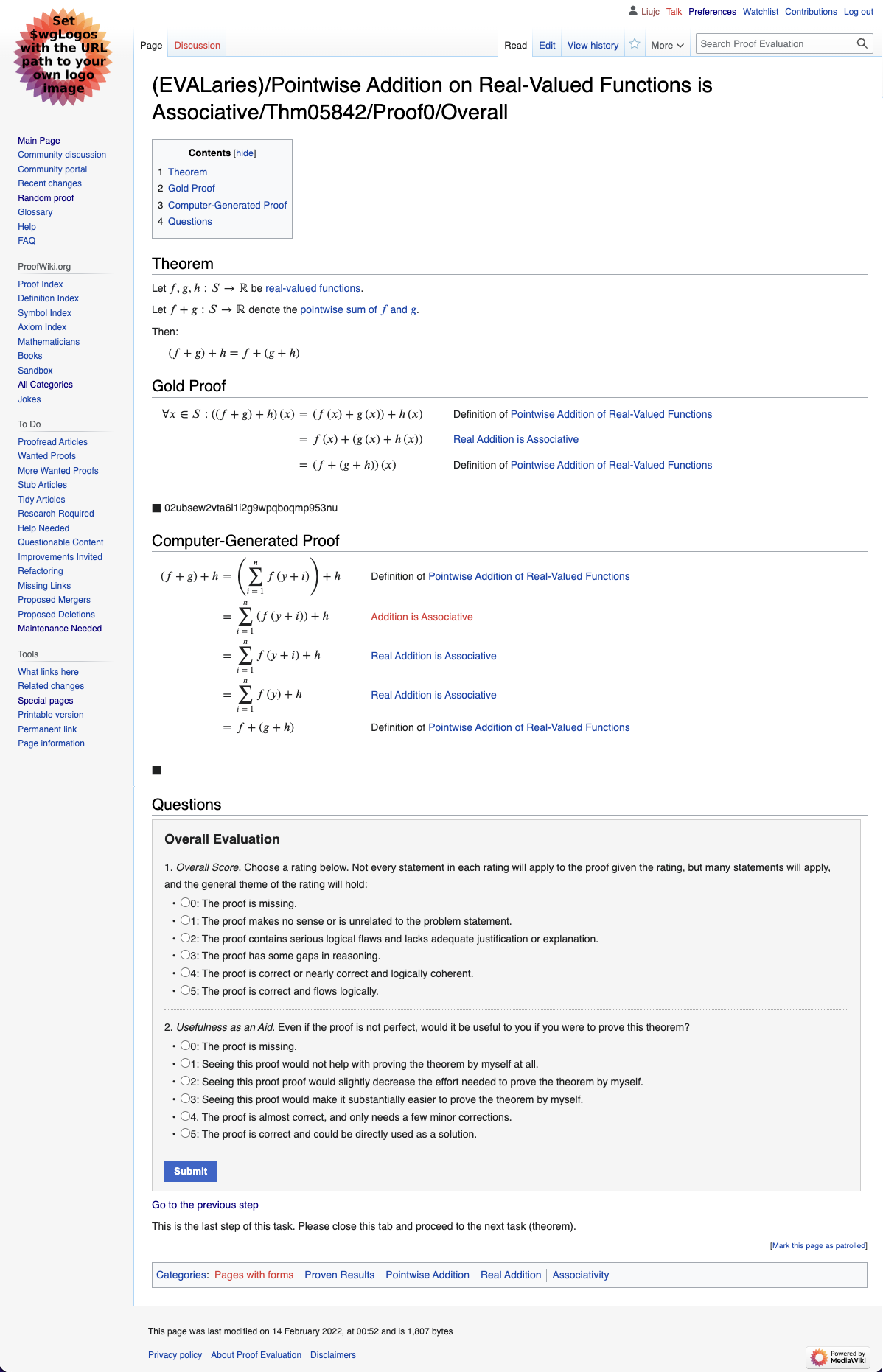}
\end{minipage}
\hspace{16pt}
\begin{minipage}{0.4 \textwidth}
    \includegraphics[width=\textwidth]{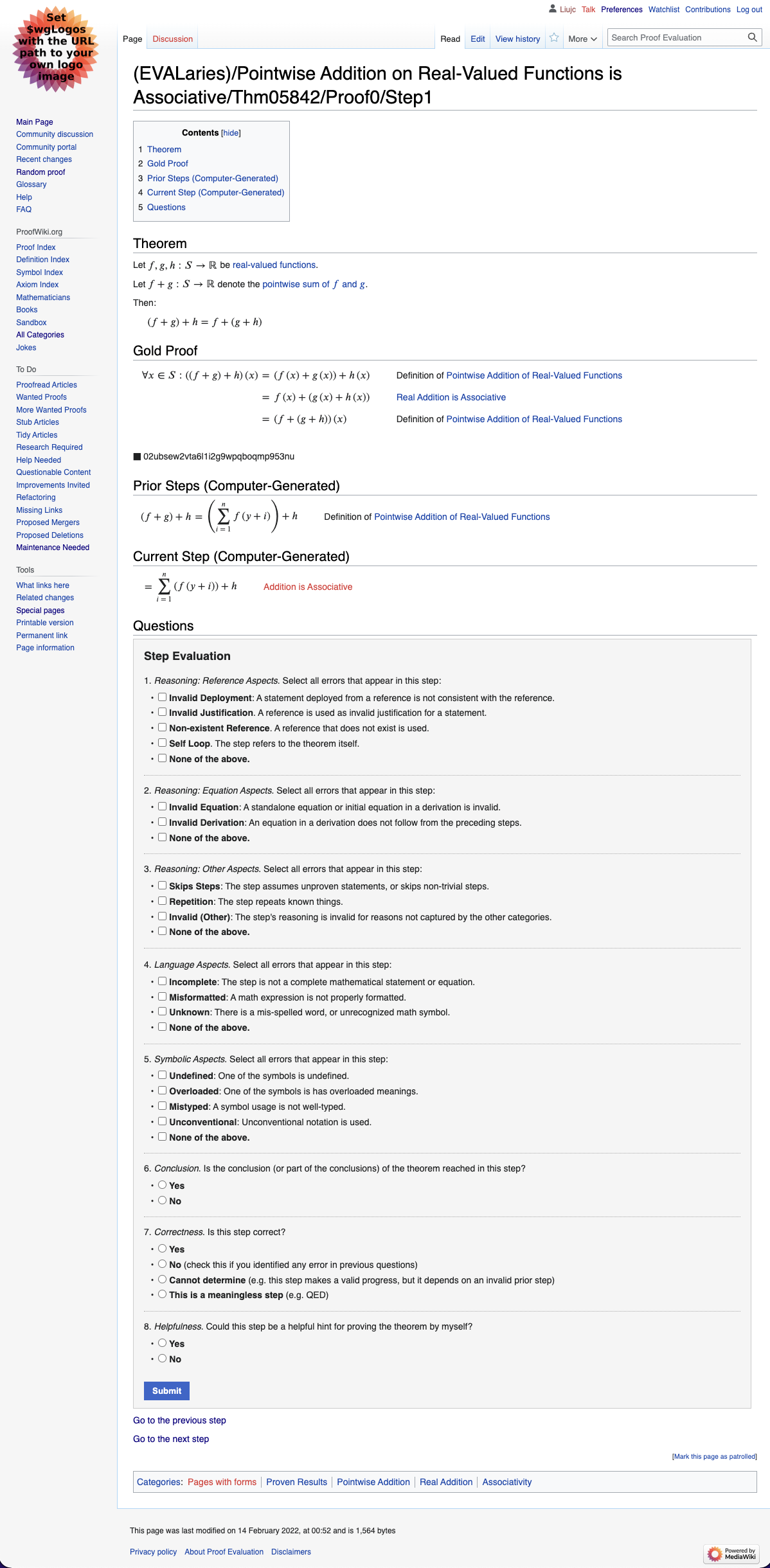}
\end{minipage}
\caption{
    Human evaluation interface.
    The first screenshot is the web console for task navigation and progress tracking.
    The next three screenshots show examples of qualification page, overall evaluation page, and step-wise evaluation page.
}
\label{fig:interface}
\end{figure}
\paragraph{Interface.}
We developed an interface that displays theorems and proofs in a rendered, human-readable format and collects annotations.
The interface is built on MediaWiki\footnote{\url{https://www.mediawiki.org}}, which also powers the ProofWiki website\footnote{\url{https://www.proofwiki.org}}.
We also developed a web console that helps human annotators navigate annotation tasks and track progress.
\autoref{fig:interface} shows screenshots of the interface.

\paragraph{Payment.}
Human annotators are paid based on the number of tasks they complete.
Each task is worth $(\$1.0 + \text{\#steps} \times \$0.4)$.
We pay each annotator an additional \$40 for attending the group session.
Annotators are guaranteed a minimal rate of \$20/hour.
The human evaluation costs approximately \$5,000.

\paragraph{Ethics review.}
The human evaluation study is approved by 
University of Washington 
under IRB STUDY00014751.
Consent was obtained from each human annotator by signing a consent form via DocuSign prior to the beginning of study.
The IRB approval letter and a template of the consent form can be found in the supplementary materials.
Minimal personally identifiable information (PII) was collected, and removed prior to any data analysis.

\subsection{Full results}
\label{sec:full_results}

\begin{table}
\footnotesize
\centering
\begin{tabular}{l c c c c | c}
\toprule
\textbf{Model} & GPT-3 & $\textsc{NP}_{\textsc{Retrieve}}$ & \textsc{NP} & $\textsc{NP}_{++}$ & \textsc{NP} \\
\textbf{Task} & Full-proof & Full-proof & Full-proof & Full-proof & Next-step \\
\bottomrule \toprule
\multicolumn{6}{c}{\textsc{Overall Evaluation} (0-5 scale)} \\
Samples & 90 & 88 & 90 & 92 & -- \\
\midrule
\cmdOverallCorrect[Correctness] ($\uparrow$) & 1.94 & 2.49 & 2.41 & \textbf{2.68} & -- \\
\cmdOverallUseful[Usefulness] ($\uparrow$) & 1.80 & 2.34 & 2.43 & \textbf{2.75} & -- \\
\bottomrule \toprule
\multicolumn{6}{c}{\textsc{Step-wise Evaluation} (\%)} \\
Samples & 802 & 727 & 654 & 466 & 665 \\
\midrule
\cmdStepCorrect[Correctness] ($\uparrow$) & 28.18 & 33.56 & 26.30 & \textbf{35.41} & 42.86 \\
\cmdStepUseful[Usefulness] ($\uparrow$) & 25.69 & 41.54 & 39.60 & \textbf{46.57} & 51.43 \\
\midrule
\cmdRefErrs[Reasoning: Reference Errors] ($\downarrow$) & 30.92 & \textbf{23.52} & 25.84 & 23.61 & 19.70 \\
\quad \cmdDeploy[Invalid Deployment] & 14.71 & \textbf{13.48} & 18.04 & 15.24 & 13.68 \\
\quad \cmdJustify[Invalid Justification] & 17.96 & 13.62 & 13.30 & \textbf{10.30} & 9.62 \\
\quad \cmdHalluc[Hallucinated Ref.] & 4.61 & \textbf{1.10} & 1.38 & 1.29 & 1.05 \\
\quad \cmdLoop[Self Loop] & 2.24 & 1.24 & \textbf{0.31} & 0.86 & 0.75 \\
\midrule
\cmdEqnErrs[Reasoning: Equation Errors] ($\downarrow$) & 32.54 & 37.55 & 35.93 & \textbf{28.54} & 26.32 \\
\quad \cmdEquation[Invalid Equation] & 15.21 & 16.23 & 12.23 & \textbf{9.44} & 12.63 \\
\quad \cmdDerivation[Invalid Derivation] & 24.56 & 27.10 & 27.37 & \textbf{21.89} & 15.64 \\
\midrule
\cmdOtherErrs[Reasoning: Other Errors] ($\downarrow$) & 40.15 & 23.66 & 25.23 & \textbf{18.45} & 19.10 \\
\quad \cmdSkip[Skips Steps] & 2.87 & 3.03 & \textbf{2.29} & 4.51 & 3.46 \\
\quad \cmdRepetition[Repetition] & 23.07 & 4.95 & 5.66 & \textbf{1.93} & 2.56 \\
\quad \cmdInvalid[Invalid (Other)] & 15.21 & 16.37 & 18.35 & \textbf{12.02} & 13.53 \\
\midrule
\cmdLangErrs[Language Errors] ($\downarrow$) & 5.61 & \textbf{4.54} & 8.41 & 5.58 & 8.57 \\
\quad \cmdIncomplete[Incomplete] & 1.62 & 2.48 & 1.99 & \textbf{1.07} & 3.76 \\
\quad \cmdMisformat[Misformatted Math] & 2.99 & \textbf{1.93} & 3.82 & 3.22 & 3.91 \\
\quad \cmdUnk[Unknown] & 1.62 & \textbf{0.69} & 3.98 & 1.72 & 2.56 \\
\midrule
\cmdSymErrs[Symbolic Errors] ($\downarrow$) & 5.24 & 6.19 & 5.35 & \textbf{3.65} & 5.86 \\
\quad \cmdUndefined[Undefined] & 1.25 & 2.06 & 1.53 & \textbf{1.07} & 2.11 \\
\quad \cmdOverloaded[Overloaded] & 2.00 & \textbf{0.41} & 0.76 & 0.43 & 0.60 \\
\quad \cmdMistyped[Mistyped] & 1.87 & 2.89 & \textbf{1.83} & 1.93 & 3.01 \\
\quad \cmdUnconventional[Unconventional] & \textbf{0.87} & 1.38 & 1.83 & 1.07 & 1.05 \\
\bottomrule
\end{tabular}
\vspace{4pt}
\caption{
    Full human evaluation results on the core test set.
    \textsc{NP} = \methodnamenospace.
    Coarse-grained error rates (e.g. \cmdRefErrs[Reasoning: Reference Errors]) are computed as the frequency of existence of \textit{any} fine-grained error under the respective bucket.
}
\label{tab:results_human}
\end{table}

\autoref{tab:results_human} shows the full results of human evaluation, including the error rates of fine-grained error types.

\subsection{Analyzing the Annotators}
\label{sec:annotators}

\begin{figure}
\centering
\begin{minipage}{0.48 \textwidth}
    \includegraphics[width=\textwidth]{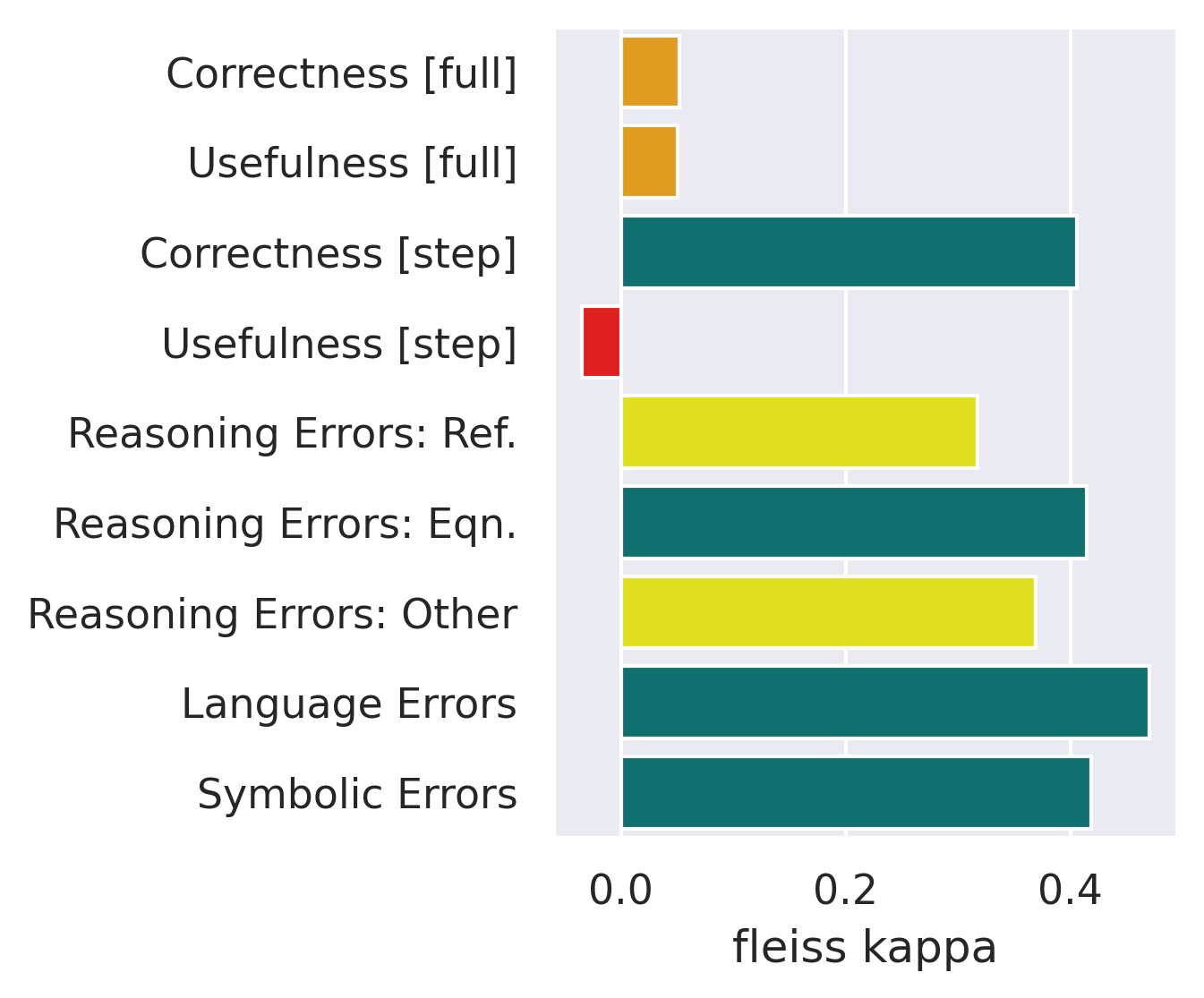}
    \caption{Inter-annotator agreement of human evaluation.}
    \label{fig:iaa_short}
\end{minipage}
\hfill
\begin{minipage}{0.48 \textwidth}
    \includegraphics[width=\textwidth]{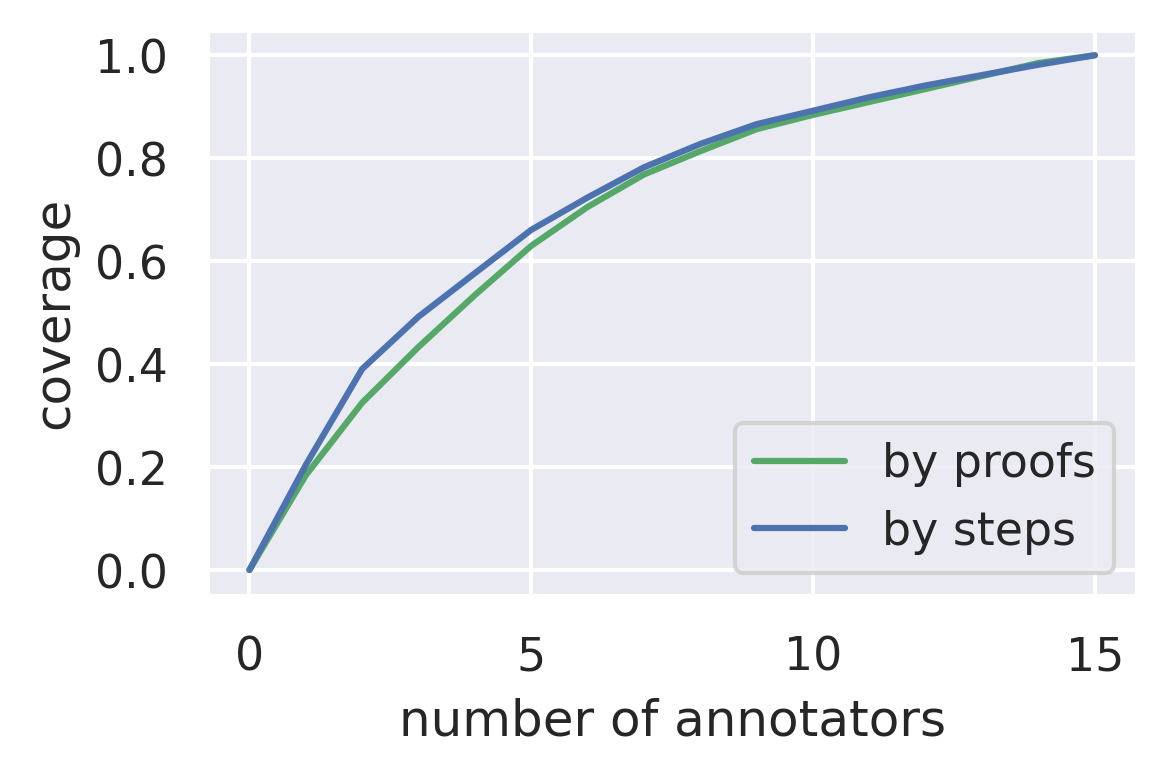}
    \caption{Source diversity of human annotations.}
    \label{fig:anno_diversity}
\end{minipage}
\label{fig:annotators}
\end{figure}

\paragraph{Inter-annotator agreement.}
We compute inter-annotator agreement using proofs in the core dev set that get an evaluation from two or more annotators.
Overall, the annotators achieved fair agreement (Fleiss kappa $\kappa = 0.24$).
The level of agreement for each evaluation question is shown in \autoref{fig:iaa_short}.
Fair to moderate agreement is reached for identifying coarse-grained error types, while the high-level questions (i.e. \textit{correctness}, \textit{usefulness}) have relatively low agreement.

\paragraph{Source diversity.}
\autoref{fig:anno_diversity} shows the largest proportion of evaluations covered by a fixed number of annotators.
The top-1 annotator contributes 20\% of the total evaluations when counting by proofs and 18\% when counting by steps.
50\% of the total evaluations is covered by roughly the top 3 or 4 annotators.
Therefore, our human evaluation results have good source diversity and do not heavily depend on a single annotator's opinion.

\section{Ethical Considerations}
\label{sec:ethical}

Our system may produce proofs of mathematical theorems that are fallacious or misleading, which may have negative impact if deployed in real educational environments.
We kindly remind potential users that our system and models are experimental, and their outputs should be interpreted critically.

\end{document}